\pdfoutput=1

\documentclass[11pt]{article}

\usepackage[final]{acl}

\usepackage{times}
\usepackage{latexsym}

\usepackage[T1]{fontenc}

\usepackage[utf8]{inputenc}

\usepackage{microtype}

\usepackage{inconsolata}

\usepackage{graphicx}

\title{iVISPAR — An Interactive Visual-Spatial Reasoning Benchmark for VLMs}

\author{
  \textbf{Julius Mayer}\thanks{\ \ Corresponding author.} \quad
  \textbf{Mohamad Ballout}\thanks{\ \ Equal contribution.} \quad
  \textbf{Serwan Jassim}\footnotemark[2] \quad \\
  \textbf{Farbod Nosrat Nezami}\footnotemark[2] \quad
  \textbf{Elia Bruni} \\
  Institute of Cognitive Science, Osnabrück University, Osnabrück, Germany \\
  \href{mailto:research@jmayer.ai}{research@jmayer.ai}
}

\usepackage{amsmath}
\usepackage{amssymb}
\usepackage{mathtools}
\usepackage{amsthm}
\usepackage{booktabs} 

\usepackage{algorithm}
\usepackage{algpseudocode}  

\usepackage{multirow}
\usepackage{tabularx}

\usepackage[most]{tcolorbox}
\usepackage{xcolor}

\usepackage{graphicx}
\usepackage{subcaption}

\usepackage{array}
\usepackage{geometry}
\usepackage[table]{xcolor}
\usepackage{colortbl}

\usepackage{enumitem}          

\setlist{nosep,leftmargin=*}   
\definecolor{custom_blue}{RGB}{202, 240, 248}
\definecolor{custom_green}{RGB}{199, 249, 204}
\definecolor{custom_yellow}{RGB}{254, 252, 233}
\definecolor{custom_gray}{RGB}{236 , 236 , 236}
\definecolor{custom_light_blue}{RGB}{245, 249, 252}

\newtcolorbox{colorbox_header}[3][]{
  colback=custom_light_blue,
  colframe=black,
  title={#2},        
  colbacktitle=#3,   
  coltitle=black,
  arc=2mm,
  boxrule=0.3mm,
  width=\linewidth,
  enhanced jigsaw,
  left=4pt,right=4pt,top=4pt,bottom=4pt,
  fonttitle=\bfseries,
  #1 
}
\newtcolorbox{colorbox_no_header}[1][]{
  colback=custom_light_blue,
  colframe=black,
  arc=2mm,
  boxrule=0.3mm,
  width=\linewidth,
  enhanced jigsaw,
  left=4pt,right=4pt,top=4pt,bottom=4pt,
  fonttitle=\bfseries,
  #1 
}

\begin{document}
\maketitle
\begin{abstract}
Vision-Language Models (VLMs) are known to struggle with spatial reasoning and visual alignment. To help overcome these limitations, we introduce iVISPAR, an interactive multimodal benchmark designed to evaluate the spatial reasoning capabilities of VLMs acting as agents. \mbox{iVISPAR} is based on a variant of the sliding tile puzzle—a classic problem that demands logical planning, spatial awareness, and multi-step reasoning. The benchmark supports visual 3D, 2D, and text-based input modalities, enabling comprehensive assessments of VLMs' planning and reasoning skills. We evaluate a broad suite of state-of-the-art open-source and closed-source VLMs, comparing their performance while also providing optimal path solutions and a human baseline to assess the task's complexity and feasibility for humans. Results indicate that while VLMs perform better on 2D tasks compared to 3D or text-based settings, they struggle with complex spatial configurations and consistently fall short of human performance, illustrating the persistent challenge of visual alignment. This underscores critical gaps in current VLM capabilities, highlighting their limitations in achieving human-level cognition. Project website: \url{https://microcosm.ai/ivispar}.
\end{abstract}


\begin{figure}[ht]
   \vskip 0.2in
    \begin{center}
        \centerline{
            \begin{minipage}{\linewidth}
                \centering
                \includegraphics[width=\linewidth]{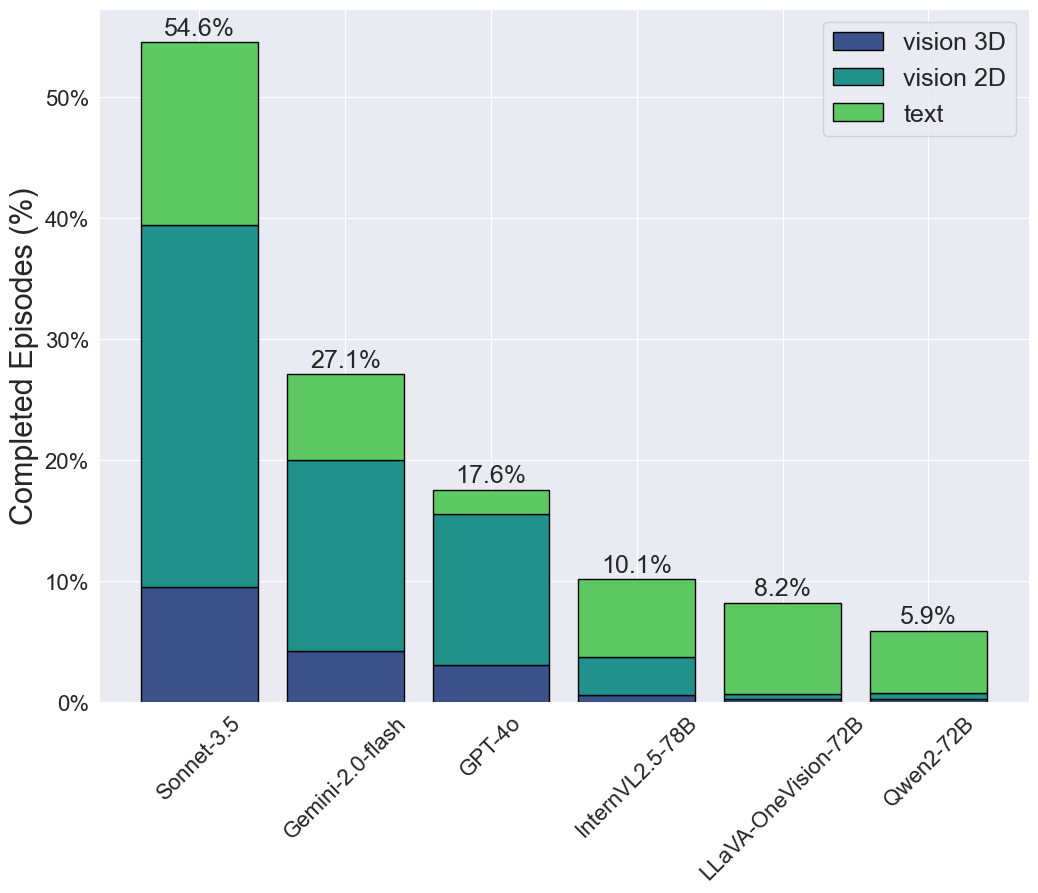}
            \end{minipage}
        }
        \caption{VLMs' success rates of completed games over 900 episodes across vision 3D, vision 2D, and text.}
        \label{fig:stacked_bar_plot_win_percentage}
    \end{center}
    \vskip -0.2in
\end{figure}

\section{Introduction}
\label{introduction}
The rapid advancement of Vision-Language Models (VLMs) has spurred significant debate regarding their capacity to achieve human-level cognition. These models are increasingly deployed as general reasoning systems capable of addressing complex problems across diverse domains, with applications extending into dynamic, real-world scenarios such as physical agent-based tasks and planning \cite{DBLP:journals/corr/abs-2406-14852, xi2023rise, zeng2023large}. However, critical gaps persist in their spatial reasoning and visual alignment capabilities, areas essential for understanding, interpreting, and manipulating objects and their spatial relationships \cite{DBLP:conf/emnlp/KamathHC23a, DBLP:journals/corr/abs-2405-17247, DBLP:journals/corr/abs-2411-00238}.

Spatial reasoning, a foundational aspect of problem-solving, navigation, and interaction with the physical world, requires models to bridge vision and cognition by interpreting visual information to understand spatial arrangements. Tasks such as mentally rotating shapes, predicting object movement, and recognizing patterns exemplify the importance of visual-spatial reasoning. Despite these critical requirements, progress in VLMs has been hampered by evaluation benchmarks that fail to capture the dynamic and multi-step complexity of real-world spatial reasoning. Existing benchmarks predominantly rely on static, text- or image-based setups that often oversimplify spatial contexts, focusing on 2D environments without interactivity or dynamic problem-solving capabilities. This limitation perpetuates a lack of meaningful progress in visual-spatial reasoning within more realistic 3D environments.

\textbf{Contributions.}\hspace{1em} To bridge this gap, we introduce iVISPAR (Interactive Visual-Spatial Reasoning), a novel benchmark designed to systematically evaluate VLMs as agents in dynamic 3D environments. iVISPAR is built around the sliding tile puzzle, a well-established problem in developmental psychology that demands logical planning, spatial awareness, and multi-step problem-solving. As part of our contributions, we introduce the Sliding Geom Puzzle, a variant that replaces traditional numbered tiles with geometric objects distinguished by their color and shape, adding an additional layer of visual reasoning.

Notably, iVISPAR is grounded in a well-studied, formalized problem with access to optimal solutions, ensuring a robust framework for evaluation.\footnote{The formalization is achieved through the adaptation of the sequential generalized sliding-tile puzzle, as described in the Appendix~\ref{slidingtilepuzzle}. Optimal solutions are computed using the A* algorithm, detailed in Section \ref{sec:baselines}.} The benchmark supports scalable task complexity by adjusting factors such as board size, the number of tiles, and solution paths, ranging from simple configurations to NP-complete challenges that surpass baseline human performance.

Leveraging a prompt-based API, iVISPAR enables VLMs to interact with a simulated environment through an iterative action-perception loop. Experimentation results demonstrate that while state-of-the-art VLMs can handle basic spatial reasoning tasks, they face significant difficulties with more complex scenarios, especially in 3D environments. Evaluating models in such 3D settings is essential, as they more closely mirror the spatial complexity of real-world environments. By contrasting their performance against optimal solutions and human baselines, we highlight the persistent gap between current VLM capabilities and human-level spatial reasoning.

Our contributions are threefold: (i) a novel interactive benchmark that systematically evaluates visual-spatial reasoning in VLMs; (ii) a scalable task design rooted in a formalized problem with optimal solutions; and (iii) empirical insights into the strengths and limitations of VLMs across varying task complexities and modalities. iVISPAR lays the foundation for advancing VLM research toward overcoming critical gaps in reasoning and alignment capabilities.



\begin{figure*}[ht]
    \vskip 0.2in
    \begin{center}
        \centerline{%
            \includegraphics[width=\textwidth]{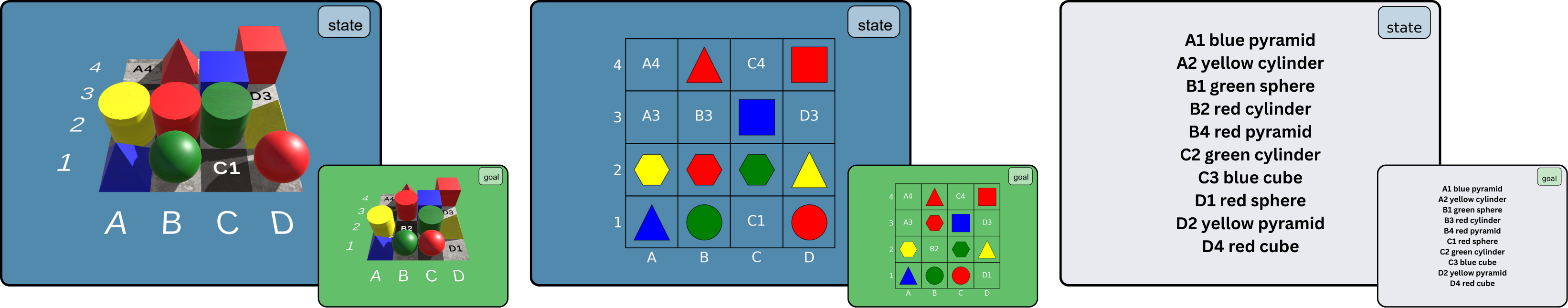}%
        }
        \caption{Example of VLMs' observations for a state (blue) and the goal (green) at each step during an episode of the Sliding Geom Puzzle environment, on a 4$\times$4 board with 10 geoms and an optimal path length of 2. Left to right, each tested modality: vision 3D, vision 2D, and text-based representation. For more examples, see Appendix \ref{sec:obs_scaling}}
        \label{fig:SGP_states}
    \end{center}
    \vskip -0.2in
\end{figure*}


\section{Related work}
\label{relatedwork}

\subsection{Spatial Reasoning Benchmarks}
\label{sec:rw_ai}
Physical understanding in interactive agents has long been studied through simulation-based benchmarks~\cite{li_muep_2024, mecattaf_little_2024, jassim_grasp_2024, wang_are_2025, hu_look_2023, zhao_embodied-r_2025, guruprasad_benchmarking_2024, su_actplan-1k_2024, feng_reflective_2025}, although many of these frameworks are not directly suited for VLM evaluation due to limited language interfaces, low task fidelity, or demanding simulation requirements.
Several datasets targeting visual reasoning have been applied to deep learning models~\cite{johnson_clevr_2016, li_super-clevr_2023}, but they do not support interactive planning or action execution by language agents.
Other works have explored similar setups using geometric object games, primarily in the context of language game learning with deep learning agents~\cite{wang_learning_2016, kuhnle_shapeworld_2017}; related efforts such as Sliding Puzzles Gym and PUZZLES~\cite{oliveira_sliding_2024, estermann_puzzles_2024} have been proposed as RL benchmarks, but lack the language interface and fine-grained 3D problem generation introduced in our setting.

\subsection{Spatial Reasoning in LLMs}
\label{sec:rw_llm}
Even though Large Language Models (LLMs) are primarily trained via next-token prediction on textual corpora, their capacity for spatial reasoning have attracted recent attention~\cite{abdou_can_2021, patel_mapping_2021}.
LLMs have also been explored as agents for spatial planning~\cite{bohnet_exploring_2024}, path planning~\cite{aghzal_can_2024}, and spatial path generation~\cite{rizvi_sparc_2024} in purely textual or symbolic environments. Several recent studies have examined whether LLMs implicitly encode spatial structures and geometric reasoning, ranging from digital twin generation via symbolic rules~\cite{wang_simbench_2024}, to textual spatial question answering in diverse settings~\cite{mirzaee_spartqa_2021}, and evaluations across grid, ring, and tree topologies~\cite{yamada_evaluating_2024}.

\subsection{Spatial Reasoning in VLMs}
\label{sec:rw_vlm}
Visual reasoning has emerged as a key focus in evaluating VLMs, with growing interest in their capacity to interpret spatial relationships and object configurations~\cite{zhang_multimodal_2024, rajabi_towards_2024, roberts_smart_2024, campbell_understanding_2025}; concurrently, several studies have examined the degree to which these models align visual inputs with linguistic representations~\cite{merullo_linearly_2023, ilharco_probing_2021}. Recent advancements in VLMs have prompted a surge in evaluations, yet most studies primarily rely on visual question-answering tests~\cite{liu_visual_2023, rajabi_gsr-bench_2024, wang_is_2024, cheng_spatialrgpt_2024, tang_sparkle_2024, duan_advancing_2025, wang_3d_aware_2023, kamath_whats_2023}.
Beyond static evaluations, a growing body of work explores the use of VLMs and foundation models as interactive agents within simulated environments, where they are tasked with manipulating objects, navigating spaces, or executing spatial instructions in grounded contexts~\cite{wu_vsp_2024, li_muep_2024, mecattaf_little_2024, jassim_grasp_2024, wang_are_2025, su_actplan-1k_2024}. This includes applications in embodied AI and robotics, where VLMs are increasingly integrated into control loops to support visuomotor reasoning and spatial decision-making~\cite{hu_look_2023, zhao_embodied-r_2025, guruprasad_benchmarking_2024, feng_reflective_2025}.

In this context, we present iVISPAR, an interactive multimodal benchmark designed to evaluate the spatial reasoning capabilities of VLMs acting as agents.




\section{The iVISPAR Benchmark}
\label{ivisparbenchmark}
iVISPAR\footnote{All source code is available under MIT license at: \\
\url{https://github.com/SharkyBamboozle/iVISPAR}} is an interactive, multimodal puzzle simulator that presents agents with a board state in one of three input modalities: a 3D rendered image, a 2D top-down view, or a text-based representation (see Figure~\ref{fig:SGP_states}). By rendering scenes in 3D space, iVISPAR offers a more realistic depiction of spatial environments compared to traditional 2D grid visualizations and enables systematic comparisons across modalities. Agents interact with the board by issuing natural language commands through a text-based API to apply actions to the board (see Figure~\ref{fig:workflow}). iVISPAR supports procedural generation of puzzle instances with finely controlled parameters, allowing for a scalable dataset of tasks with adjustable complexity across many spatial properties, and benchmarking performance with multiple baseline models.



\begin{figure*}[ht]
   \vskip 0.2in
    \begin{center}
        \centerline{
            \begin{minipage}{\linewidth}
                \centering
                \includegraphics[width=\linewidth]{imgs/other/game_flow.png}
            \end{minipage}
        }
        \caption{Depiction of the interaction flow between VLM agents and the iVISPAR simulator with a progression through an episode with the shortest path solution of 4 steps being solved by prompted actions from a VLM agent. For a full example of an episode progression, see Appendix \ref{sec:episode_progression}.}
        \label{fig:workflow}
    \end{center}
    \vskip -0.2in
\end{figure*}

\subsection{Sliding Geom Puzzle}
\label{slidinggeompuzzle}
A central environment in iVISPAR is the Sliding Geom Puzzle (SGP), a reimagining of the classic sliding tile puzzle (see Appendix~\ref{slidingtilepuzzle}). Instead of numbered tiles, SGP uses geometric objects (geoms) uniquely defined by combinations of color and shape, increasing visual-spatial complexity and enhancing task scalability. This design shift requires models to interpret object features rather than follow numerical sequences, mirroring real-world spatial reasoning where items are distinguished by appearance, size, or structure. The task draws inspiration from physical scenarios such as organizing items, assembling structures, or packing, promoting a more authentic evaluation of real-world spatial capabilities.


\subsection{Game dynamics}
\label{sec:dynamics}
The objective is to rearrange the pieces on the board by moving them over free spaces to match a given goal configuration. In each episode, agents receive observations of the start and goal states (see Figure~\ref{fig:SGP_states}), accompanied by task instructions (see Appendix~\ref{system_prompts}). Agents apply move actions to geoms by referencing their unique color and shape combination and specifying the direction of intended movement. Geoms can be moved in cardinal directions (\textit{LEFT, RIGHT, UP, DOWN}), with actions formatted as "move $<$color$>$ $<$shape$>$ $<$direction$>$”:

\begin{verbatim}
"move blue sphere right"
\end{verbatim}

Actions are validated and applied if legal, with agents receiving updated board states regardless of the action’s success after each move command. Effective
and ineffective actions both result in valid new board states
but, respectively, decrease or increase the path length to
the goal state. Invalid moves, such as occupied destination
and out-of-bounds actions, fail to alter the board state, as
do illegal commands, which violate the instructed action
format. This action-perception loop repeats until the goal state is achieved or a step limit is reached. Due to limited context windows, VLM agents receive task instructions at each time step. A sample agent-environment interaction is provided in Appendix \ref{interaction}.

\subsection{Observation Spaces}
\label{sec:observations}
Agents observe a combination of the current board state and the goal state. Additionally, they can receive a sequence of past state-action pairs, determined by the size of the configured context window. 
Images for 3D observations are presented from an angled top-down perspective and may include partially occluded objects, whereas 2D observations follow a graph-like layout with fully visible elements. Both may optionally include embedded, text-based chess-style coordinate labels as spatial cues along the outer edge of the grid board as well as on free tiles. In 2D observations, shapes are mapped consistently from their 3D counterparts to preserve object identity across modalities. Images can also be marked with an embedded text label and a colored background to differentiate between past (grey), current (blue), and goal state (green). Figure~\ref{fig:SGP_states} shows 3D vision (left) and 2D vision (middle) for the active state (top) and the goal state (bottom). The text-based representation encodes past, active, and goal states directly in the prompt string supplied to the agent. Agents receive the list of geoms in the order of board coordinates. A visualization of the text-based active (top) and goal states (bottom) is shown in Figure~\ref{fig:SGP_states} (right). This modality does not rely on images.

\subsection{Complexity Scalability}
\label{sec:complexity_scalability}
The GSTP is a well-known NP-hard problem due to the need for multi-step planning across a constrained grid \cite{gozon2024computing}. SGP inherits this complexity but introduces greater flexibility in scaling difficulty without altering the game’s core mechanics. This flexibility provides more degrees of freedom, making the task more tractable for VLM agents. Key scaling factors include board size, number of objects, object variability, length of the shortest path solution, and the geom interference factor (see Appendix \ref{sec:obs_scaling}). The shortest path solution for all episode configurations is calculated using the A* algorithm \cite{hart1968formal}, as detailed in Appendix \ref{sec:optimal_agent}. The interference factor denotes the extent to which objects obstruct one another’s optimal paths, increasing the global solution length beyond the cumulative Manhattan distances of individual paths. This interference can create configurations with short optimal paths but increased planning requirements, significantly raising the problem's difficulty. Available geometric shapes include [“cube,” “pyramid,” “sphere,” “cylinder,” “cone,” “prism”], with colors freely selectable by referencing RGB values. Agents must navigate combinatorial complexity by matching shapes and colors, promoting spatial strategies over the sequential patterns seen in numerical tile puzzles. Episode configurations are generated procedurally, requiring models to generalize across puzzle instances. Human and algorithmic benchmarks for these experiments are detailed in Section \ref{sec:baselines}.


\section{Experiments}

Performance of VLMs is tested for the SGP to assess their capabilities in scene understanding, problem-solving, and multi-step planning within constrained environments.

\subsection{Dataset Generation}
Experiments were conducted on a dataset 
of SGPs on a fixed board size to $4\times4$: smaller grids (e.g., $3\times3$) collapse many spatial-relation cases, while larger ones ($\geq 5\times5$) dilute object visibility without yielding further complexity benefits. Performance is assessed by varying complexity across two parameters: the number of objects (2-11) and the shortest path length (2-11). Configurations maintain a geom interference factor of 0, ensuring the shortest path equals the cumulative Manhattan distance. Initial experiments indicated that VLM agents faced significant challenges at higher task complexities. Three episodes are sampled for each complexity level, producing a dataset of 300 diverse board configurations. The set of geom properties consists of four shapes, sphere, pyramid, cube, and cylinder, and four colors, red, green, blue, and yellow, resulting in 16 unique combinations. VLM agents are tested on the same dataset for each modality, resulting in 900 episodes for each model.


\subsection{Baselines}
\label{sec:baselines}
To contextualize agent performance and provide upper and lower bounds, we establish four baselines encompassing human and AI agents. 

\textbf{Human performance} was evaluated with 30 participants using a web app GUI of the SGP, where participants interacted by prompting text commands over a command line, mirroring the interaction method of VLM agents. Baselines were provided for the 3D vision modality on the same dataset as the VLM agents. 

\textbf{AI baselines} were introduced for two agents: an optimal agent executing shortest path solutions computed by A* \cite{hart1968formal}, and a random agent performing uninformed but valid actions uniformly sampled from those leading to new board states. Algorithms for the AI agents are detailed in Appendix \ref{sec:algorithms}.


\begin{figure*}[ht]
    \vskip 0.2in
    \begin{center}
        \includegraphics[width=\linewidth]{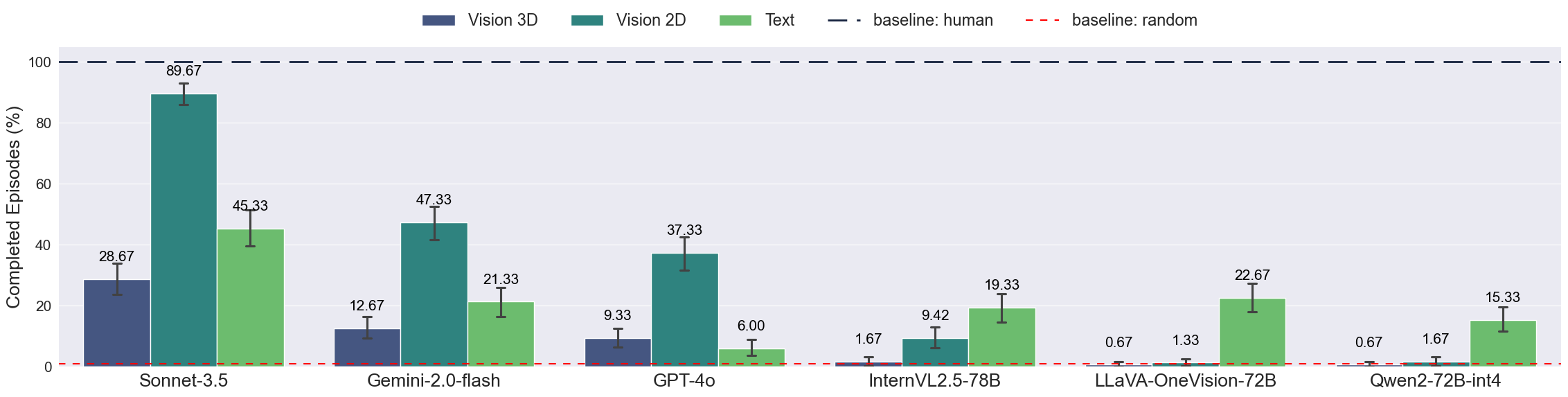}
        \vskip 0.2in
        \includegraphics[width=\linewidth]{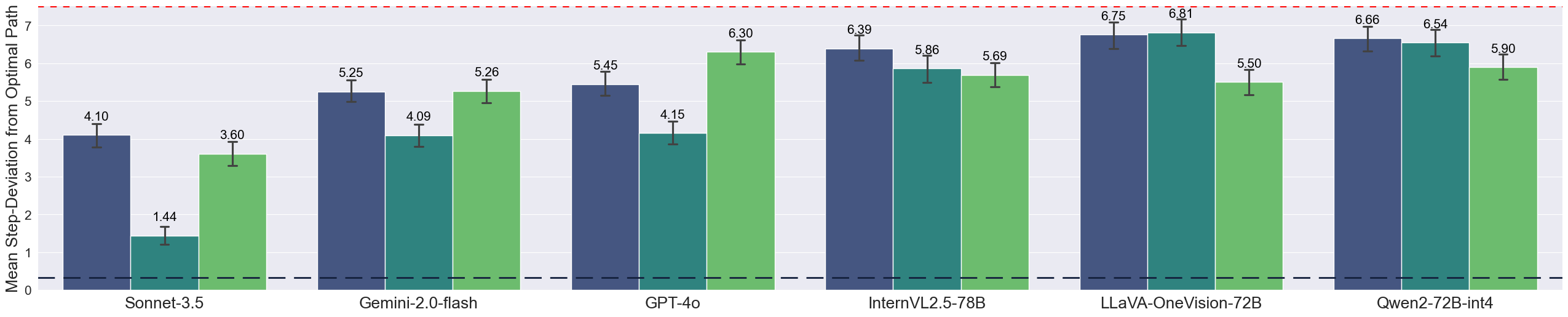}
        \caption{VLM evaluation on 900 episodes per model across all three modalities, with 95\% confidence intervals. Baseline comparisons for human performance and random moves are shown. Top: VLMs' success rates of episodes completed with higher values denoting better performance. Bottom: VLMs' mean step deviation from the optimal path with lower values denoting better performance. Full numerical results are provided in Appendix~\ref{sec:detailed_results}}
        \label{fig:results_bar_plot}
    \end{center}
    \vskip -0.2in
\end{figure*}

\subsection{Models}
\label{sec:models}
We evaluate a selection of open- and closed-source VLMs that scored high on OpenCompass\footnote{OpenCompass Official Rankings: \url{https://rank.opencompass.org.cn/leaderboard-multimodal}} and which support multi-image inputs and a minimum context length of 800 tokens. Selected models are: Sonnet-3.5 \cite{claudesonnet35}, Gemini-2.0-flash \cite{gemini20flash}, GPT-4o \cite{gpt4o}, InternVL2.5-78B \cite{chen2024expanding}, LLaVA-OneVision-72B \cite{li2024llava}, Qwen2-72B \cite{Qwen2VL}. For closed-source models, we rely on the official APIs and for open-source models, on the publicly available checkpoints. We use a temperature of 1.0, top-p of 0.95, and top-k of 50 for all open-source models. An overview of all models and their details can be found in the Appendix~\ref{app:models}.

\subsection{Context-Aware Zero-Shot Reasoning}
The models employ Chain-of-Thought (CoT) reasoning \cite{wei2022chain} to break down complex problems into smaller sub-tasks, enhancing accuracy and interpretability (Appendix \ref{interaction}). We constrain VLMs’ context windows to the past two steps, incorporating state representations alongside the model’s action responses. This approach prioritizes extracting maximum value from limited experience to preserve the models’ sequential coherence and minimize computational overhead. Operating within this context-aware zero-shot reasoning framework, the models interpret task requirements without examples,  drawing exclusively from pretrained knowledge, task instructions, and limited past interactions.


\subsection{Instruction Prompts}
\label{sec:prompts}
We avoided prompt engineering for any single model; the chosen template is the same for all systems and contains only the minimal information needed. Fixing one validated template provides a consistent basis for comparison and makes the benchmark easily reproducible. The visual and text prompts are isomorphic: the image placeholder is the only difference, so no modality receives extra hints. Our human-baseline study likewise found the final wording easy to follow. This supports our aim of testing spatial-reasoning ability itself, without relying on prompt engineering, so we use one clear, uniform template for all models.

\subsection{Evaluation}
\label{sec:eval}
Agent performance is evaluated through two primary metrics: the fraction of solved environments and mean step-deviation from the optimal path 

\textbf{Mean step-deviation from optimal path} measures the deviation from optimal behavior during problem-solving. At each step \( t \), the shortest path solution from the current board state to the goal, computed by A*, is used to assess efficiency. Formally,
\[
R(t) = d(s_t, s^*) - \bigl[d(s_0, s^*) - t\bigr].
\] 
where \( d(s, s^*) \) denotes the shortest path length from state \( s \) to the goal \( s^* \). 
This metric quantifies how much further the agent is from the goal compared to an optimal agent after the same number of steps. 
A regret value of zero indicates that the agent follows an optimal trajectory, while positive regret reflects inefficiencies or unnecessary detours. 
By capturing performance even in unsolved environments, this approach provides insights into agent behavior under varying complexities.

To gain deeper insights, we analyze the most common error patterns exhibited by agents. This allows us to identify model weaknesses, recurring failure cases, and patterns of suboptimal decision-making.

\subsection{Auxiliary Task}
\label{sec:auxiliary_task}
Additionally, we evaluate the models’ ability to infer and represent board states from visual input across all 300 episodes. Given an image and accompanying instructions, each model is tasked with predicting the corresponding board configuration in text form, using the same format as the textual representation shown in Figure~\ref{fig:SGP_states}. This auxiliary task further enriches our understanding of the models’ behavior and their capacity to interpret spatial information from visual inputs.

To analyze this task, we frame the comparison between the true and predicted board states as a set matching problem, solved using the Hungarian algorithm. A match is defined as any pair of geoms sharing at least either color or shape. Geoms that share neither are considered missed (if only present in the true state) or hallucinated (if only present in the prediction). Matched geoms may still contain mismatches in coordinates, color, or shape. Predicted elements that cannot be parsed into valid geom triplets are counted as format errors.


\begin{figure*}[ht]
    \vskip 0.2in
    \begin{center}
        \includegraphics[width=\linewidth]{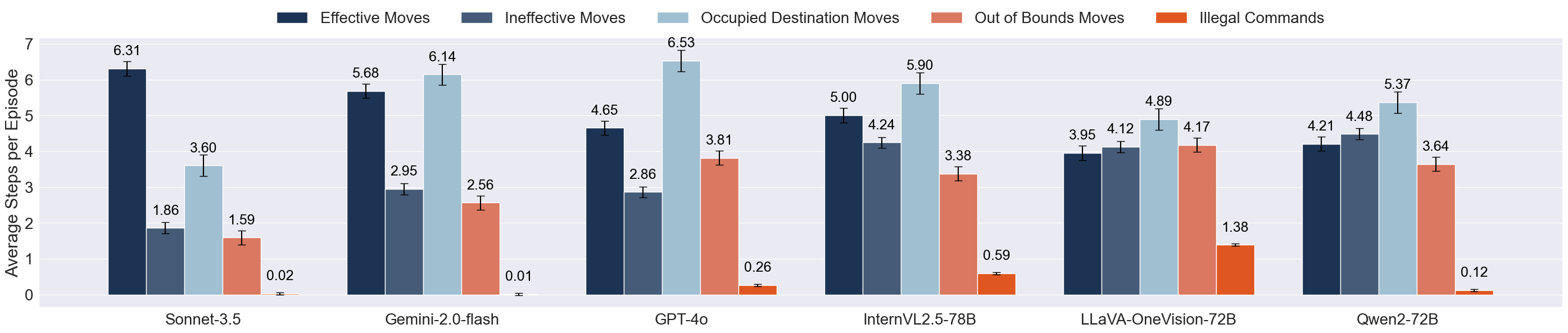}
        \vskip 0.2in
        \includegraphics[width=\linewidth]{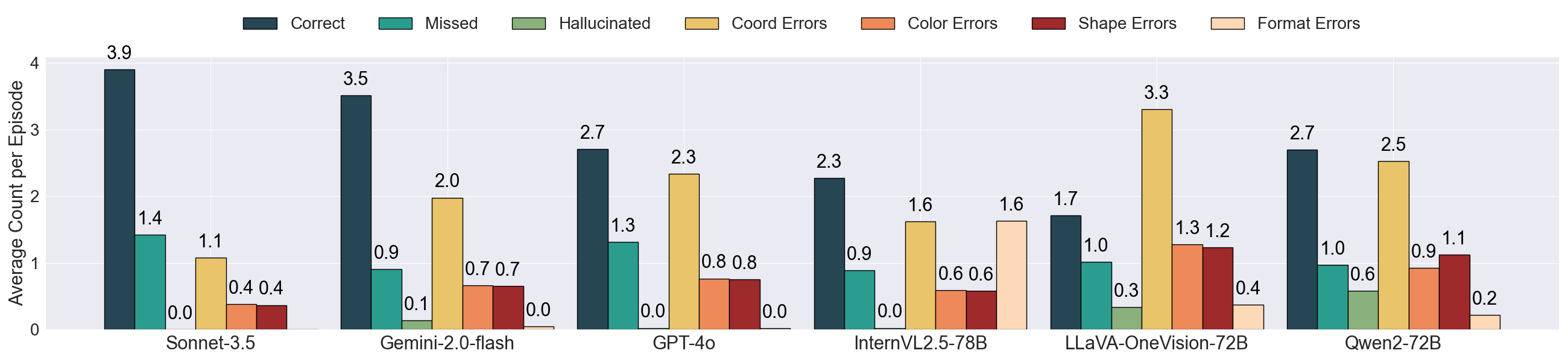}
        \caption{Error patterns showing average action counts per episode during SGP interaction (top) and average geoms per episode for the board state inference auxiliary task (bottom), both averaged across modalities (see Sections~\ref{results} and~\ref{sec:auxiliary_task}), each aggregated across modalities. Full numerical results are provided in Appendix~\ref{sec:detailed_results}.}
        \label{fig:move_validity_double}
    \end{center}
    \vskip -0.2in
\end{figure*} 

\section{Results}
\label{results}


\renewcommand{\thefootnote}{\fnsymbol{footnote}}
\footnotetext[1]{Videos of exemplary agents' interactions with iVISPAR are available at:
\url{https://youtu.be/Djis_xkgtW8}.}
\renewcommand{\thefootnote}{\arabic{footnote}}

We evaluated the spatial reasoning capabilities of VLMs in our SGP environment on 3D vision and compared it to 2D vision and text-based modalities across 300 episodes each (see Figure \ref{fig:results_bar_plot}). To standardize gameplay, the number of actions per episode was capped at 20.

\textbf{Success rates:} The percentage of episodes completed and the mean deviations of steps from the optimal path were measured for each modality and compared to human performance as well as random actions (Figure \ref{fig:results_bar_plot}).

\textbf{Action classification:} We classified actions based on their effects on the board and calculated their average occurrence per episode to provide insights into the challenges VLMs face in efficiently completing episodes (see Figure \ref{fig:move_validity_double} top). Effective and ineffective actions both result in valid new board states but, respectively, decrease or increase the path length to the goal state. Invalid moves, such as occupied destination and out-of-bounds actions, while illegal commands break the instructed action format, all of which leave the board state unchanged. 

\textbf{Auxiliary Task:} For the board state inference task, we evaluate the number of geoms that were correctly inferred, missed, hallucinated, or contained a mismatch in coordinates, color, or shape. Format errors denote cases where the output failed to follow the expected structure (Figure \ref{fig:move_validity_double}, bottom).

\textbf{Complexity scales:} We evaluated the cumulative performance of VLMs across the three modalities using two complexity scales, the shortest path length required to solve an episode and the number of geoms on the board. Longer shortest paths demand a broader global planning horizon and consistent goal-directed progress, while higher geom counts require efficient local planning to optimize rearrangement order and manage free spaces. Figure \ref{fig:cumulative_heatmap} illustrates the performance of VLMs in 100 combinations of complexity, highlighting the average minimal distance to the goal state in 20 steps.

\section{Discussion}
\label{discussion}


\begin{figure*}[ht]
    \vskip 0.2in
    \begin{center}
        \includegraphics[width=0.49\linewidth]{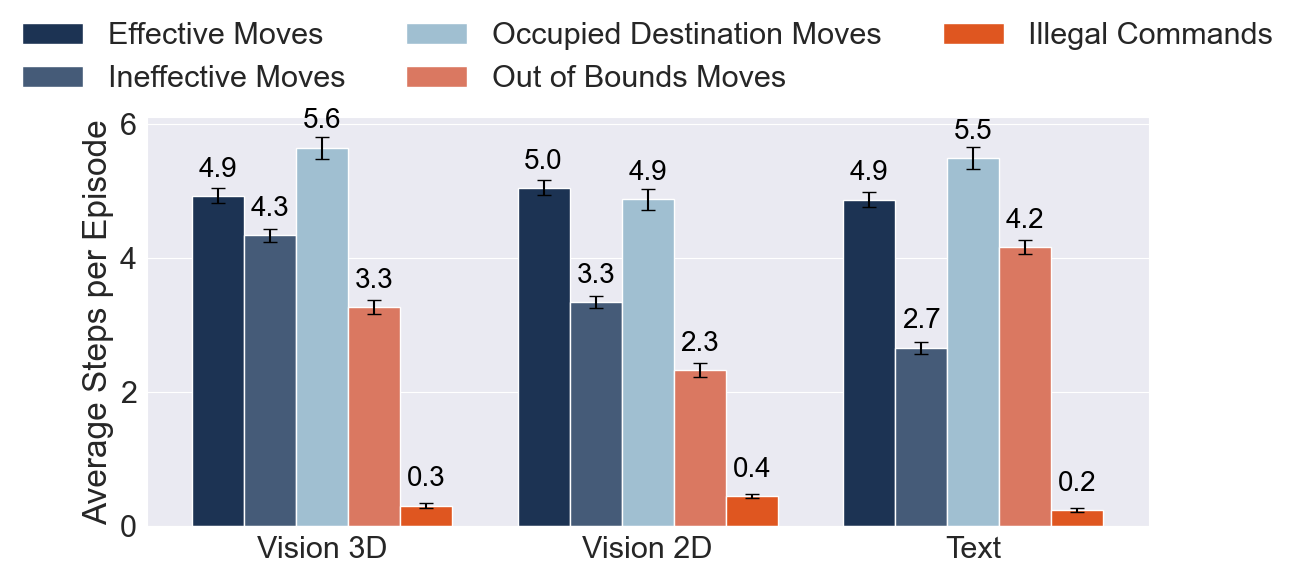}%
        \hfill
        \includegraphics[width=0.49\linewidth]{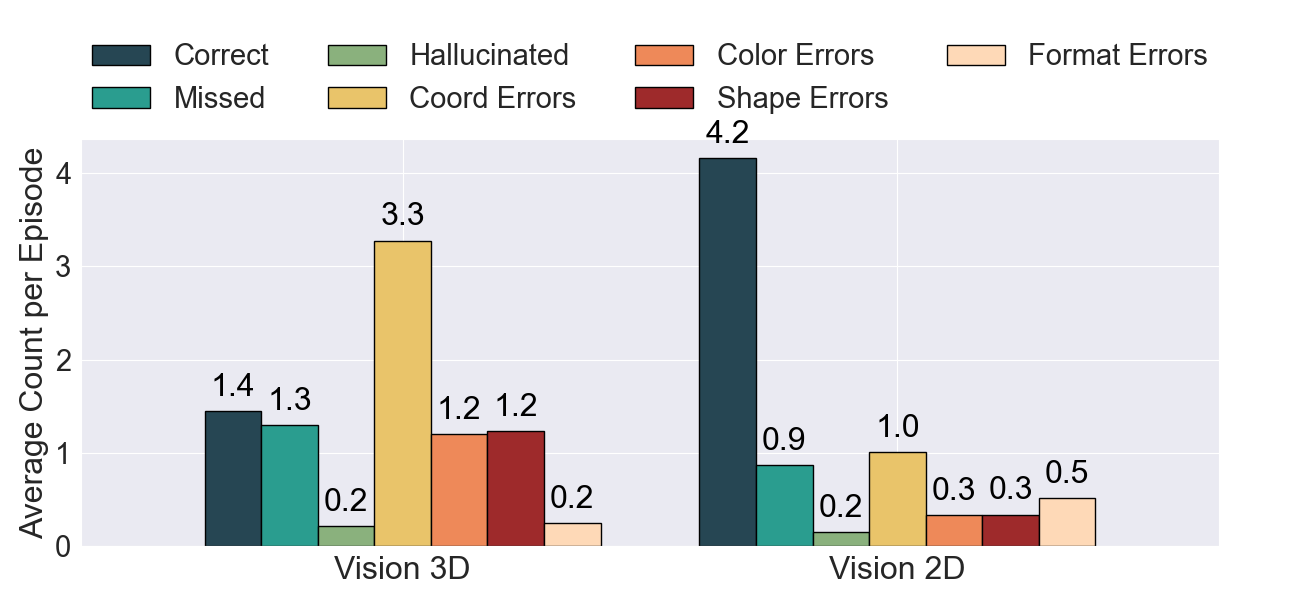}%
        \caption{Error patterns showing average action counts per episode during SGP interaction (left; see Section~\ref{results}) and average geoms per episode for the board state inference auxiliary task (right; see Section~\ref{sec:auxiliary_task}), shown per modality and aggregated across agents. Full numerical results are provided in Appendix~\ref{sec:detailed_results}.}
        \label{fig:modalities_only}
    \end{center}
    \vskip -0.2in
\end{figure*}

\subsection{Model Performance}
\label{sec:EMNLP_vlm_perform}
All models show basic task understanding and spatial reasoning, progressing toward the goal state (see Figure~\ref{fig:results_bar_plot}). Performance, however, varies widely. Closed-source models outperform open-source ones: Sonnet-3.5 achieves the highest success rate at 89.7\% in the 2D visual modality, followed by Gemini-2.0-Flash and GPT-4o. In contrast, open-source models such as InternVL2.5-78B, LLaVA-OneVision-72B, and Qwen2-72B perform near the random baseline. Human participants solve the tasks perfectly with near-optimal paths, setting a high benchmark.

Notably, even models solving fewer than 1\% of tasks often produce more efficient paths than a random baseline (see Figure~\ref{fig:results_bar_plot}, bottom), indicating traces of goal-directed behavior despite overall failure. These task performances are also consistent with the further analysis of the models’ error types and their accuracy in the board state inference task, which we discuss in Section~\ref{sec:EMNLP_error_analysis}.
\subsection{Error Patterns}
\label{sec:EMNLP_error_analysis}
We analyzed the types of mistakes models make during interaction with the simulator and evaluated their ability to infer board states from visual input. Overall, models rarely issue illegal commands or exhibit format errors (see Figure~\ref{fig:move_validity_double}, top and bottom), suggesting that most VLMs understand how to follow instructions and interact with the environment appropriately.

However, board state inference accuracy reveals a sharp performance drop from 2D to 3D inputs: while models correctly identify an average of 4.2 objects in 2D, this number falls to 1.4 in the 3D setting (see Figure~\ref{fig:modalities_only}, right). This is primarily due to substantial increases in coordinate prediction errors, alongside moderate rises in color, shape mismatches, and missed detections. In contrast, hallucinations and format-related issues remain largely stable across both modalities.

These findings offer a clear explanation for the weaker performance in the 3D vision condition: precise localization of objects remains a critical challenge. As illustrated in Figure~\ref{fig:move_validity_double}, this results in more ineffective moves, including frequent attempts to place objects out-of-bounds or onto already occupied cells.
\subsection{Modality Impact}
\label{sec:EMNLP_modality_influence}


\begin{figure*}[ht]
    \vskip 0.2in
    \begin{center}
        \includegraphics[width=\linewidth]{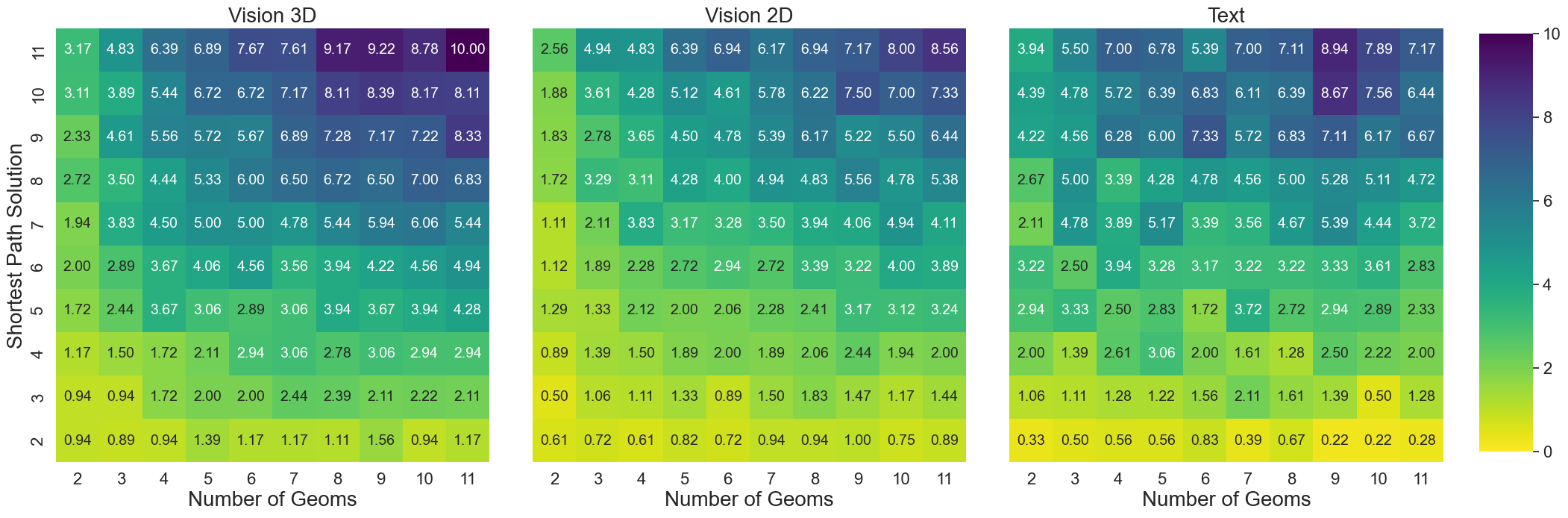}
        \vskip -0.2in
        \includegraphics[width=\linewidth]{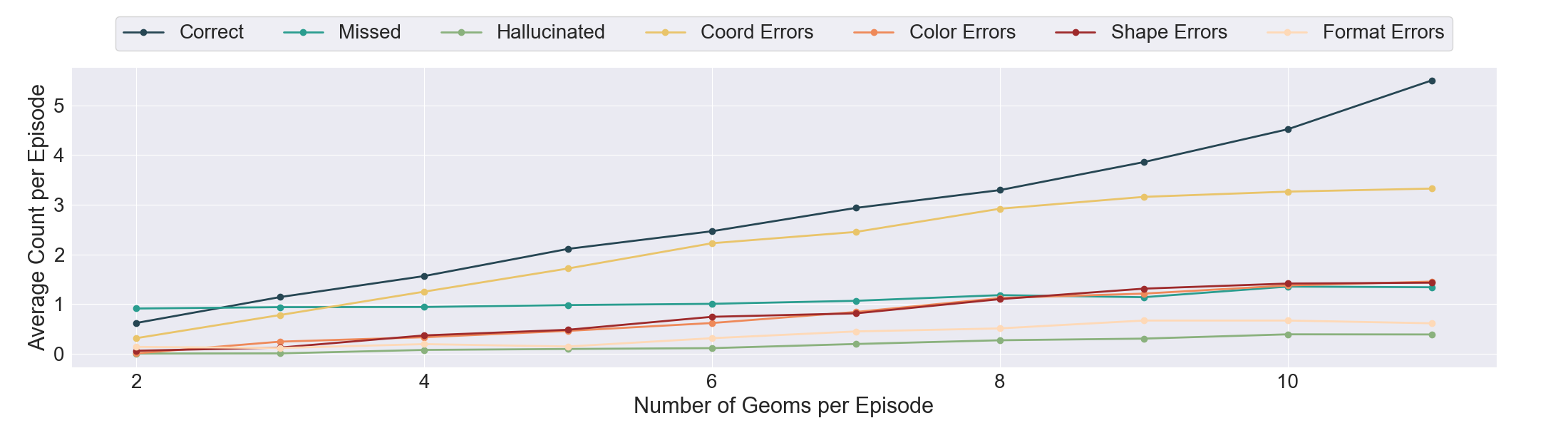}
        \caption{Cumulative graphs aggregated across agents. Top: Correlation matrix of remaining shortest-path lengths to the goal for tasks with optimal paths between 2–11 steps. Each run is capped at 20 actions, and the metric is computed at the agent’s final state, either upon reaching the goal or, if unsolved, after the 20th action. Bottom: Error types in the board state inference auxiliary task over increasing number of geoms on the board.}
        \label{fig:cumulative_heatmap}
    \end{center}
    \vskip -0.2in
\end{figure*}

Despite being evaluated on identical tasks, model performance varied substantially across input modalities (see Figure~\ref{fig:results_bar_plot}). All closed-source models (Sonnet-3.5, Gemini-2.0-flash, GPT-4o) performed best on 2D vision, followed by text, and worst on 3D vision. This suggests that these models may have undergone more training on 2D visual inputs, which are more common in spatial benchmarks. Interestingly, text input, despite posing significant challenges for humans, ranked second, indicating some robustness in linguistic reasoning. 
In contrast, open-source models (InternVL2.5, LLaVA-OneVision, Qwen2) performed poorly across the board, with near-random scores on visual inputs. Their relatively stronger performance on text tasks may reflect a reliance on superficial pattern recognition rather than grounded spatial understanding. 
As shown in Figure~\ref{fig:modalities_only} (left), error patterns for ineffective moves and collisions align with the overall performance ranking across modalities. Out-of-bounds errors are most frequent in the text condition, nearly twice as common as in 2D vision, indicating that understanding board dimensions was a primary challenge in the textual setting.
Additional results from our board state inference task further support this view, showing that models, predict more correct objects on the board in Vision 2D compared to Vision 3D (Figure~\ref{fig:modalities_only}, right).

\subsection{Complexity Scaling}
\label{sec:EMNLP_complexity_scaling}
We analyzed the correlation matrix between the number of objects on the board and the shortest path solution length to assess how different types of complexity affect model performance (see Figure~\ref{fig:cumulative_heatmap}, top). While performance consistently drops with increasing complexity in both dimensions, the heatmaps reveal modality-specific trends. Performance declines more steeply with increasing geom count (particularly in 3D), suggesting that sequential planning under visual conditions poses a major challenge. In contrast, in the text-only setting, the number of geoms seems to have little effect, with errors mostly determined by the length of the shortest path solution. This highlights limitations in spatial reference from language alone.

Data from the auxiliary task of board state inference show that, while errors to predict the coordinates of geoms on the board increase with the number of geoms on the board, other error types remain relatively stable even for a higher number of geoms on the board (see Figure~\ref{fig:cumulative_heatmap}). Format errors and the number of hallucinated geoms is overall low, mismatches with colors and shapes increasing only slightly, and surprisingly the number of missed objects stays relatively stable as well.


\section{Conclusion}
\label{conclusion}
We have introduced iVISPAR, a novel interactive multimodal benchmark designed to evaluate the spatial reasoning capabilities in 3D vision of VLMs acting as agents. The benchmark, centered on the Sliding Geom Puzzle, evaluates VLMs' abilities in logical planning, spatial awareness, and multi-step problem-solving, aiming to reflect real-world spatial reasoning. Our evaluation tested a suite of state-of-the-art open-source and closed-source VLMs on a dataset of board configurations, scaled across two levels of complexity. We compared them to baselines for human capabilities, optimal and random agents, providing insight into their performance under varying conditions.

Our findings demonstrate that VLMs struggle with spatial reasoning in 3D vision and that there are significant performance differences between the tested VLMs. While they understand the instructions and outperform random agents in simple spatial tasks, they struggle with more complex configurations and intricate problem properties. Interestingly, VLMs show stronger performance in 2D vision compared to 3D or text-based tasks. Our auxiliary board state inference task revealed that VLMs frequently miss geoms, misplace them on the board, or mismatch their colors or shapes, errors that occur more often with 3D vision input than with 2D. This suggests that visual alignment for 3D spatial reasoning continues to pose a significant challenge, underscoring persistent gaps in VLM capabilities and highlighting barriers to achieving human-level cognitive performance.

\textbf{Future Work}\hspace{.3cm} Looking ahead, we plan to expand the benchmark to incorporate additional tasks focused on scene understanding, as well as rotation and transformation challenges.

\textbf{Resources}\hspace{.3cm} For the most up-to-date results on state-of-the-art models and access to the leaderboard, please visit: \\
\url{https://microcosm.ai/ivispar}.

\section*{Acknowledgments}
\label{Aknowledgement}
This work was funded by the Deutsche Forschungsgemeinschaft (DFG, German Research Foundation) — 456666331, 321892712.

\section*{Limitations} 
\label{sec:limitations}
We restricted the context window, limiting the number of images VLMs can process. Extended image inputs often disrupt VLMs' understanding of sequential coherence and increase computational demands and API costs. This contrasts with human participants, who recall each step of an episode and draw from past experiences.

Additionally, while some models are optimized for long-context reasoning or "deep thinking," their architecture and usage patterns are ill-suited for step-wise, interactive simulations. Their per-frame API costs are disproportionately higher, making them impractical for the interaction format used in our benchmark. This also limits direct comparisons to human participants, who recall previous steps and integrate episodic knowledge more efficiently.

\section*{Impact Statement}
This paper contributes to advancements in vision-language models. While our work has potential applications in broader AI research, it does not introduce immediate ethical or societal risks beyond those already associated with the field. As our work is largely theoretical and not at a scale that could pose significant concerns, it does not raise specific risks of misuse or unintended consequences.




\newpage
\appendix
\section{Appendix}



\subsection{Episode Details}
\label{sec:episode_details}

\subsubsection{System Prompt Instructions}
\label{system_prompts}

\begin{colorbox_header}{Interactive Sliding Geom Puzzle Game}{custom_blue}
You are a highly intelligent AI solving a shape puzzle on a 4x4 grid. The board has two states: the current active state and the goal state. Your task is to generate valid actions that transform the current state into the goal state along the shortest path.

\vspace{0.5em}

\textbf{Steps:} 

(1) Analyze current state. 

(2) Compare to goal. 

(3) Check past actions. 

(4) Propose next move.
\vspace{0.5em}

\textbf{Movement Rules:} Each object occupies one tile. Objects cannot leave the grid or overlap.

\vspace{0.5em}

\textbf{Action Format:} \texttt{move <color> <shape> <direction>}

\vspace{0.5em}

Use only the following:\\
\textbf{Colors:} green, red, blue, yellow\\
\textbf{Shapes:} cube, sphere, pyramid, cylinder\\
\textbf{Directions:} up, down, left, right

\vspace{0.5em}

\textbf{Examples:} \texttt{move green cube down}, \texttt{move red pyramid left}

\vspace{0.5em}

\textbf{Important:} No coordinates. Each action must change the state. Invalid if blocked or out of bounds.

\vspace{0.5em}

\textbf{Explain Reasoning:} Before suggesting an action, explain why. End with:\\
\texttt{action: move <color> <shape> <direction>}\\
(no extra characters after \texttt{action: ...})

\vspace{0.5em}

{\raggedright \textbf{Visual Input:}\par}
\mbox{Current: \texttt{\{text\_snippet\_active\}};}
\mbox{Goal: \texttt{\{text\_snippet\_goal\}};}
\mbox{Past: \texttt{\{text\_snippet\_past\}}.}

\vspace{0.5em}

\textbf{Final Requirement:} Always end your output with:\\
\texttt{description:<your object coordinate list>}\\
Do not add 
characters after the word \texttt{description}.
\end{colorbox_header}

\begin{colorbox_header}{Board State Inference Auxiliary Task}{custom_blue}

You are a highly intelligent AI with exceptional spatial reasoning skills, and you are given the following task: 

\vspace{0.5em}

\textbf{\#\# Task Overview}

\begin{enumerate}
    \item You are provided with an input image of colored geometric objects on a 4 $\times$ 4 board.
    \item Analyze the current board state and locate the position of all objects on the board.
    \item Respond with a list of the chess-style coordinates and their objects.
\end{enumerate}

\vspace{0.5em}

\textbf{\#\# Board Overview}

The board has labeled columns, rows, and fields
\begin{itemize}[nosep,leftmargin=*]
\item Columns a–d run from left to right in the image.
\item Rows 1–4 run from bottom to top in the image.
\end{itemize}

\vspace{0.5em}

\textbf{\#\# Object Overview}

\begin{itemize}[nosep,leftmargin=*]
\item On the board are various objects, uniquely defined by their color and shape:
  \begin{itemize}[nosep,leftmargin=*]
      \item \textbf{Colors:} green, red, blue, yellow  
      \item \textbf{Shapes:} cube, sphere, pyramid, cylinder
  \end{itemize}
\end{itemize}

\vspace{0.5em}

\textbf{\#\# Solution Format}

\begin{itemize}[nosep,leftmargin=*]
    \item Start your solution with `\verb|Solution: |` and list each object in any order, separated by a comma and a single space (`, `).
    \item Your solution for each object must follow this exact format: \verb|<coordinate>| \verb|<object_color>| \verb|<object_shape>|
      \begin{itemize}[nosep,leftmargin=*]
          \item \verb|coordinate| must use a letter a–d followed by a digit 1–4.
          \item \verb|object_color| must be exactly one of: green, red, blue, yellow.  
          \item \verb|object_shape| must be exactly one of: cube, sphere, pyramid, cylinder.  
      \end{itemize}
    \item Only list coordinates that contain an object; do not mention empty squares.
    \item Do not use quotation marks or angle brackets < > in your action.
    \item Do not include any extra text, reasoning, or punctuation after the formatted list.
\end{itemize}

\vspace{0.5em}

\textbf{\#\# Example}

Solution: a3 green sphere, d1 blue cylinder, b4 yellow cube, c2 red pyramid

\vspace{0.5em}

\textbf{\#\# Validation}

\begin{itemize}[nosep,leftmargin=*]
\item No two objects share the same coordinate.
\item Every listed object uses one of the four allowed colors and shapes.
\end{itemize}
\end{colorbox_header}

\onecolumn

\subsubsection{Observations of Scaling Episode Complexity}
\label{sec:obs_scaling}

\begin{figure*}[ht]
    \vskip 0.2in
    \begin{center}
        \includegraphics[width=\linewidth]{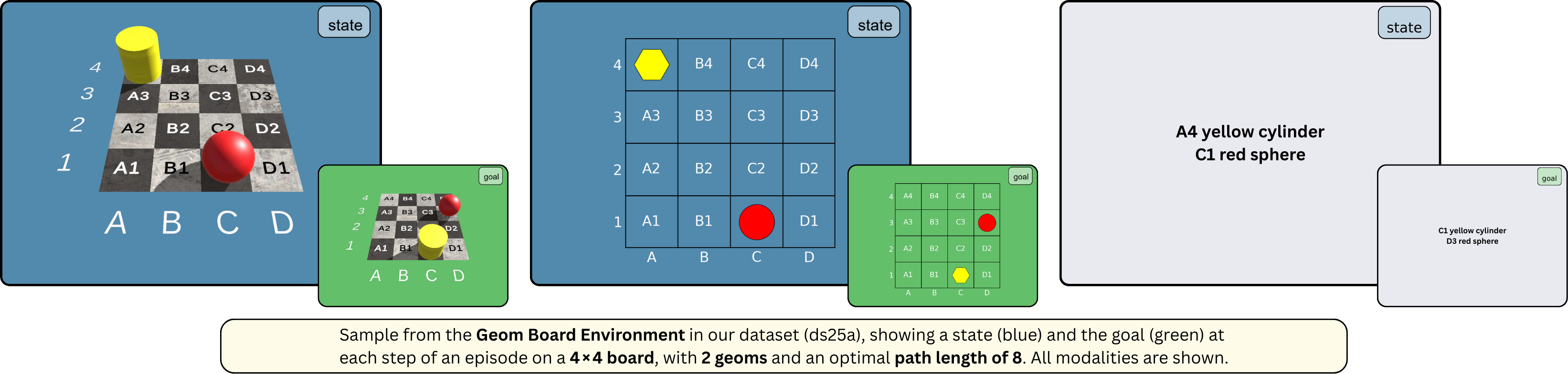}
        \vskip 0.25in
        \includegraphics[width=\linewidth]{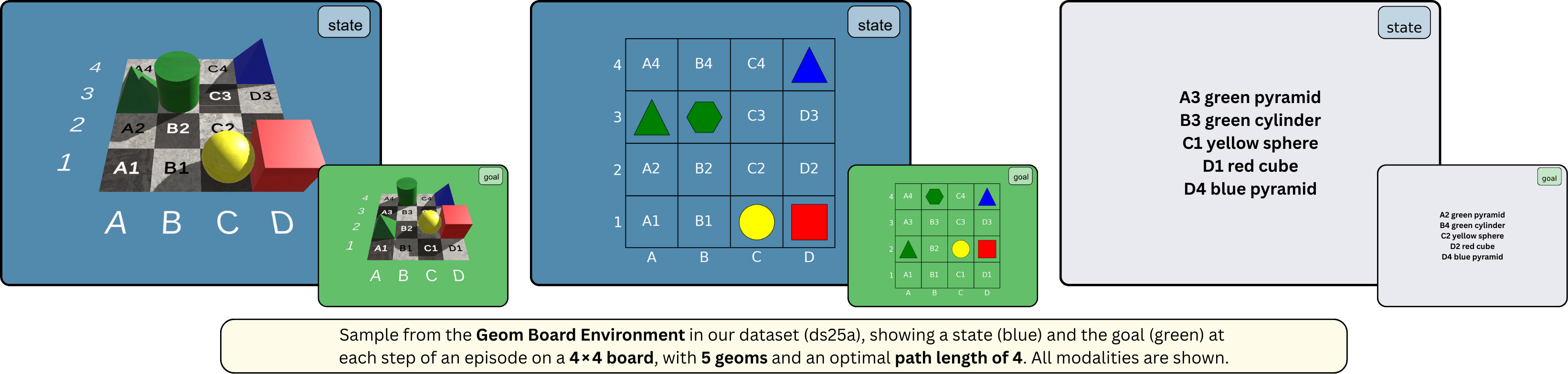}
        \vskip 0.25in
        \includegraphics[width=\linewidth]{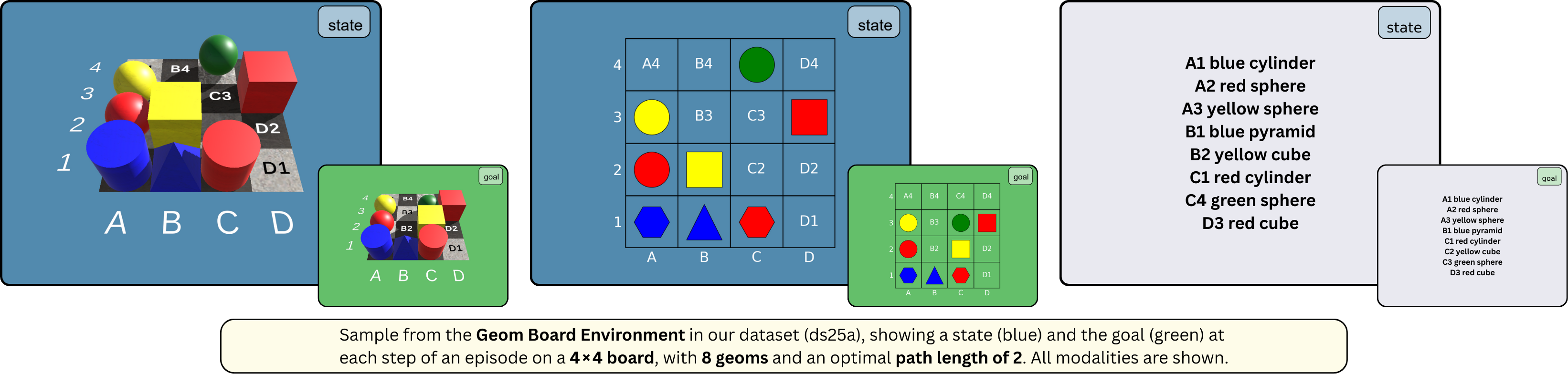}
        \vskip 0.25in
        \includegraphics[width=\linewidth]{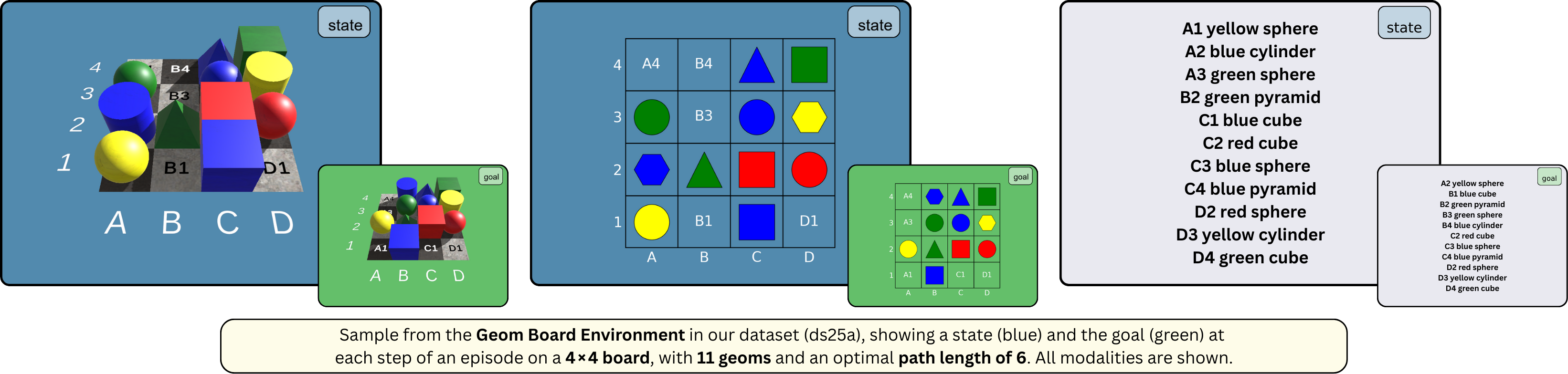}
        \caption{Examples of VLMs' observations for a state (blue) and the goal (green) at each step during an episode of the Sliding Geom Puzzle environment, on a 4$\times$4 board with 2, 5, 8 and 11 geoms and an optimal path length of 8, 4, 2, and 6 respectively. Left to right, each tested modality: vision 3D, vision 2D, and text-based representation.}
        \label{fig:observations_scaling}
    \end{center}
    \vskip -0.2in
\end{figure*}

\newpage
\subsubsection{Interaction Example}
\label{interaction}
\begin{colorbox_header}{Input Example}{custom_blue}

\begin{center} 
  \includegraphics[width=0.9\linewidth]{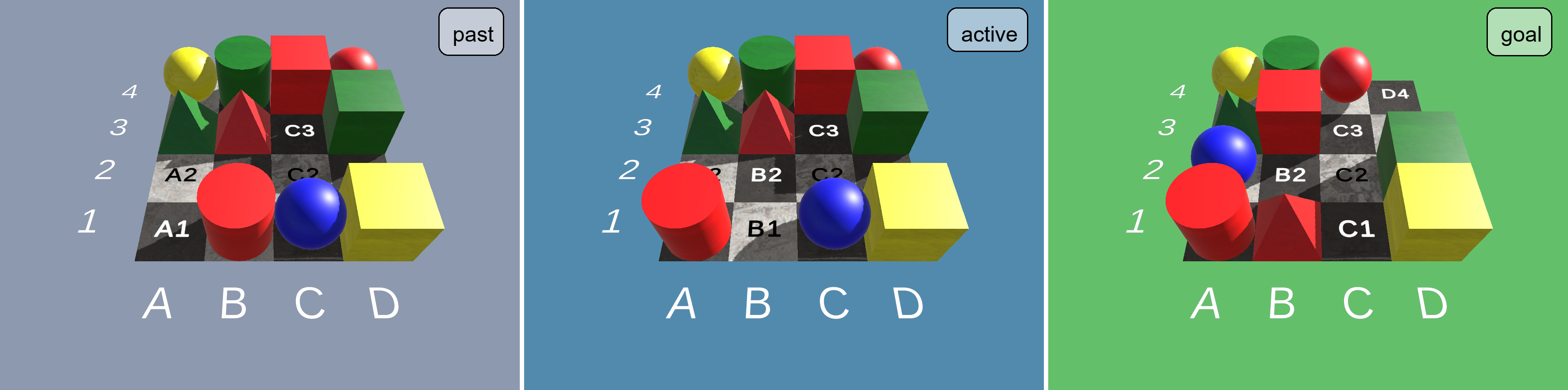}  
\end{center}

\textbf{\#\# Analyze the Images}\\
You can view your current active board state in the last image marked with the label \textit{active} in the image and a blue background. Study this image and the objects with their positions carefully.\\

Your goal is to match the goal state, shown in the image marked with the label \textit{goal} in the image and a green background. Study this image and the objects with their positions carefully.\\

\textbf{\#\# Additionally, you are provided with:}
\begin{itemize}[nosep,leftmargin=*]
\item The previous state image(s) marked with the label \textit{past} in the image and a grey background.
\item Your previous suggested action
\item Use this information by comparing it to your current active state to determine your next action.\\
\end{itemize}

\textbf{\#\# Invalid Actions:}
\begin{itemize}[nosep,leftmargin=*]
\item No Overlap: You are not allowed to position two objects in the same tile.
\item If the suggested action does not move any objects, it is invalid (e.g., blocked by another object or out of bounds).
\item Use the previous image(s) and action to understand why it failed and suggest a different action.
\end{itemize}

\vspace{.3cm}

It is of most importance you always end your response with this exact format:

\begin{center}
\verb|action: move <object color> <object shape> <direction>|
\end{center}

where you replace \verb|<object color> <object shape> <direction>| with the valid move action based on your reasoning and do not add any characters after your action.

\end{colorbox_header}
\begin{colorbox_header}{Output Example}{custom_green}

To reach the goal state I need to compare the current active state with the goal state.

\vspace{.3cm}

\textbf{Current active state}
\begin{itemize}[nosep,leftmargin=*]
\item In the current active state the green sphere is on top blocking movements of objects below it.
\item The yellow sphere is on the bottom row B1.
\end{itemize}

\vspace{.3cm}

\textbf{Goal state}
\begin{itemize}[nosep,leftmargin=*]
\item In the goal state the green sphere is two steps down in column B.
\item The red cylinder is in the bottom left, two steps down.
\item The yellow sphere is in position B2, two steps up.
\end{itemize}

\vspace{.3cm}

Analyzing the images I see that the yellow spheres path is blocked by the green cylinder I’ll first move another object to clear the way. Moving the blue cube rightward will help clear the blocked pathway for the green cylinder.

\vspace{.3cm}

\textbf{Action:} Move the blue cube right

\end{colorbox_header}

\newpage
\subsubsection{Episode Progression Example}
\label{sec:episode_progression}

\begin{figure*}[ht]
    \centering
    \vskip 0.2in

    \begin{subfigure}{0.32\linewidth}
        \includegraphics[width=\linewidth]{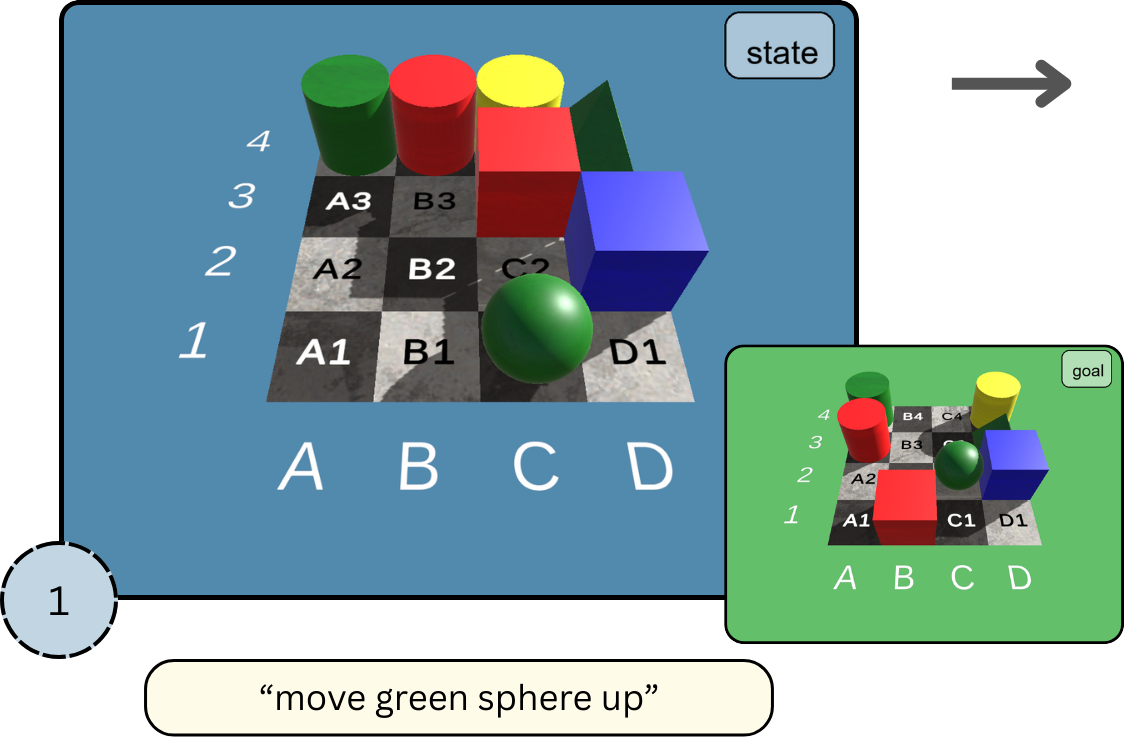}
    \end{subfigure}
    \hfill
    \begin{subfigure}{0.32\linewidth}
        \includegraphics[width=\linewidth]{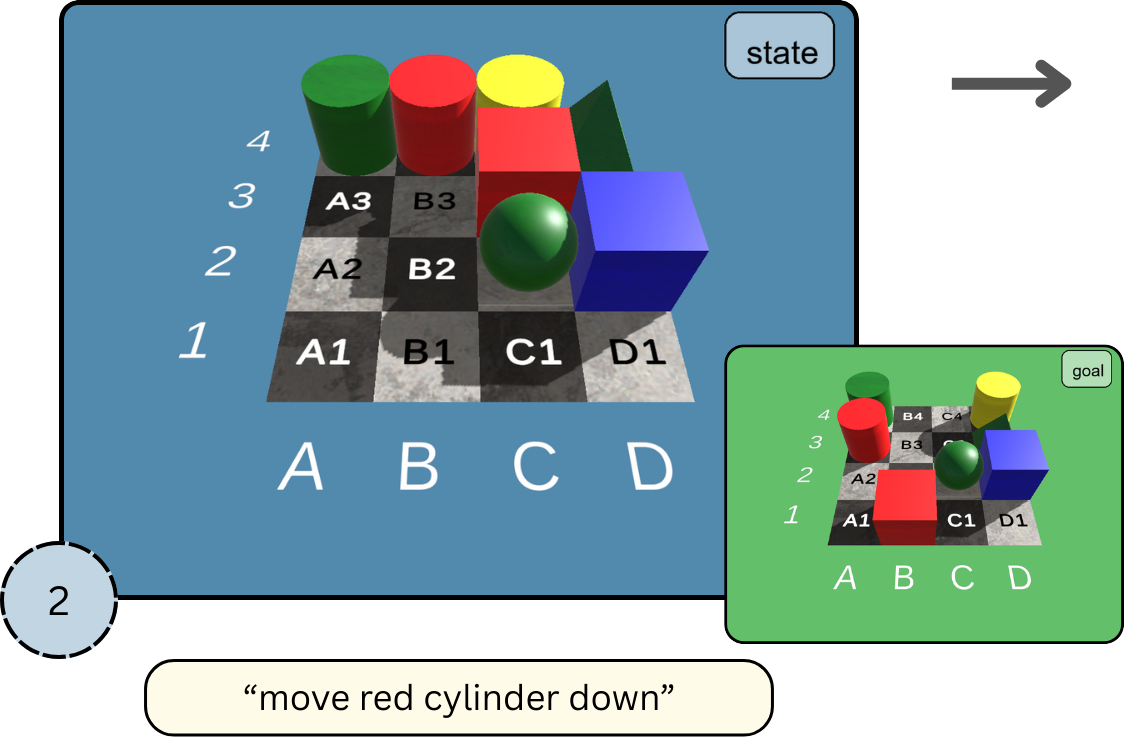}
    \end{subfigure}
    \hfill
    \begin{subfigure}{0.32\linewidth}
        \includegraphics[width=\linewidth]{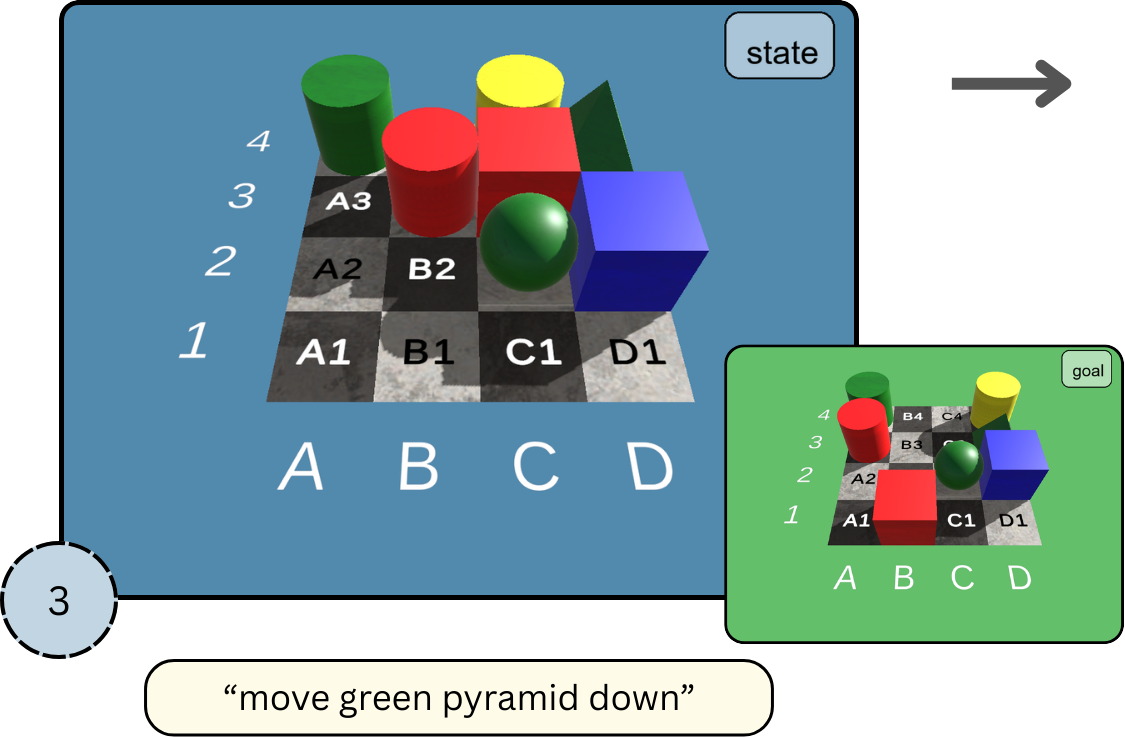}
    \end{subfigure}

    \vskip 0.3in

    \begin{subfigure}{0.32\linewidth}
        \includegraphics[width=\linewidth]{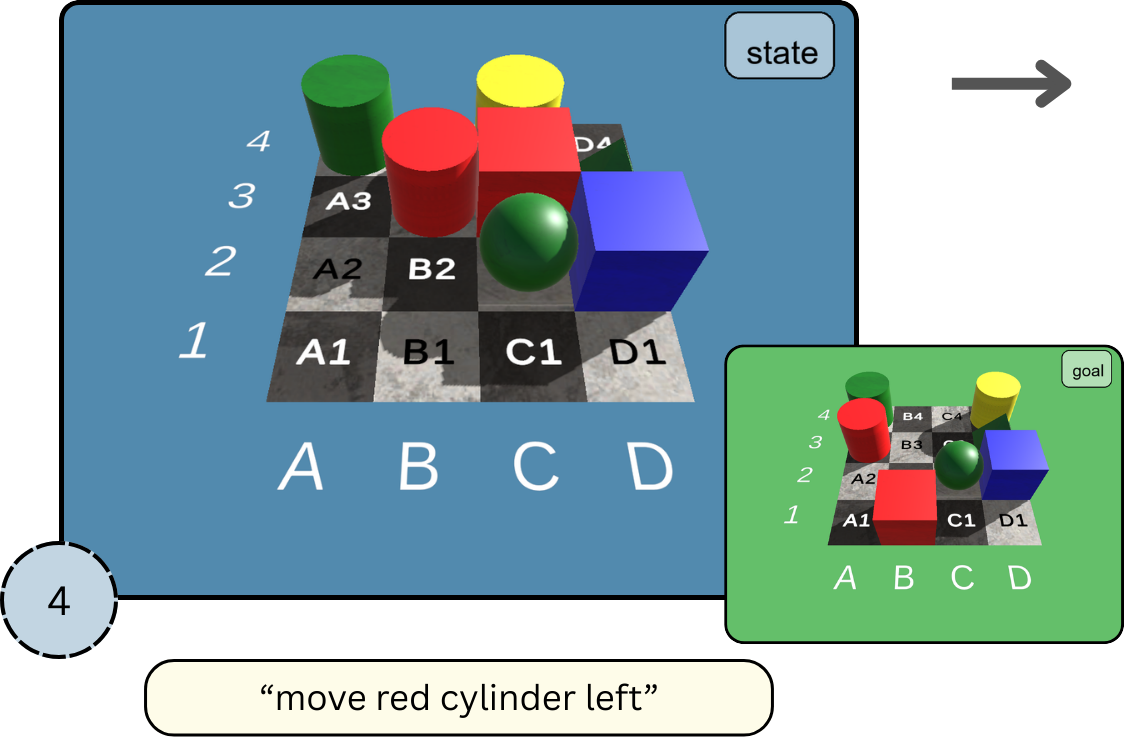}
    \end{subfigure}
    \hfill
    \begin{subfigure}{0.32\linewidth}
        \includegraphics[width=\linewidth]{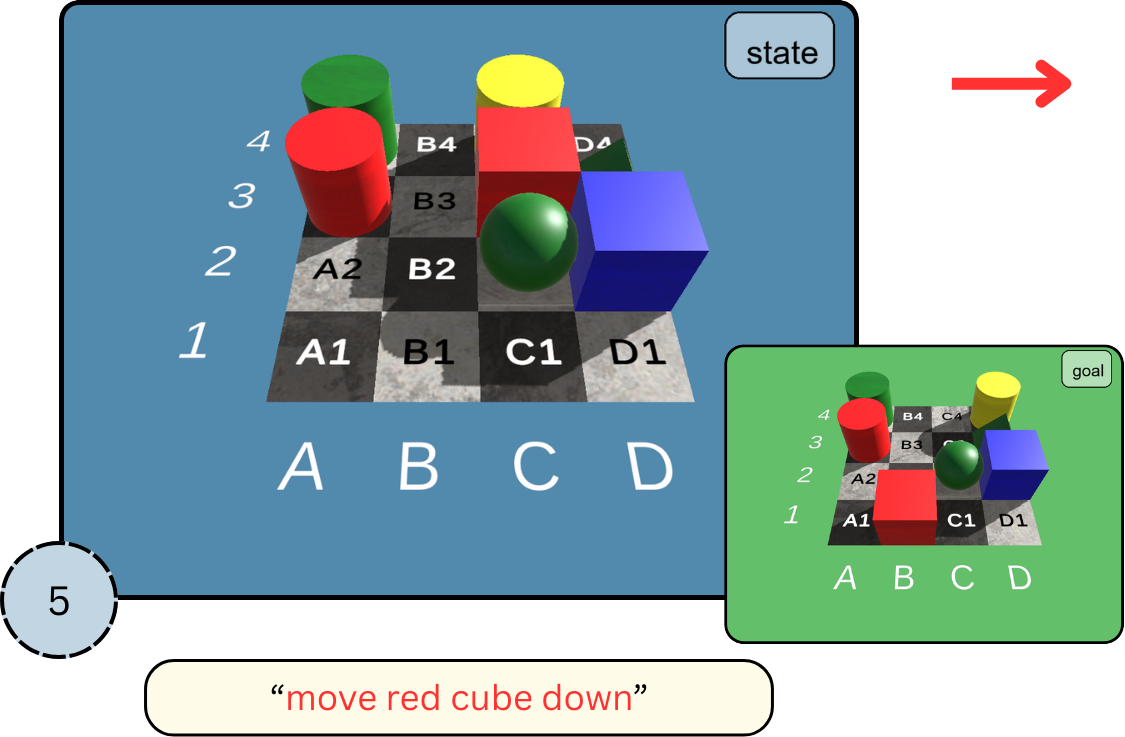}
    \end{subfigure}
    \hfill
    \begin{subfigure}{0.32\linewidth}
        \includegraphics[width=\linewidth]{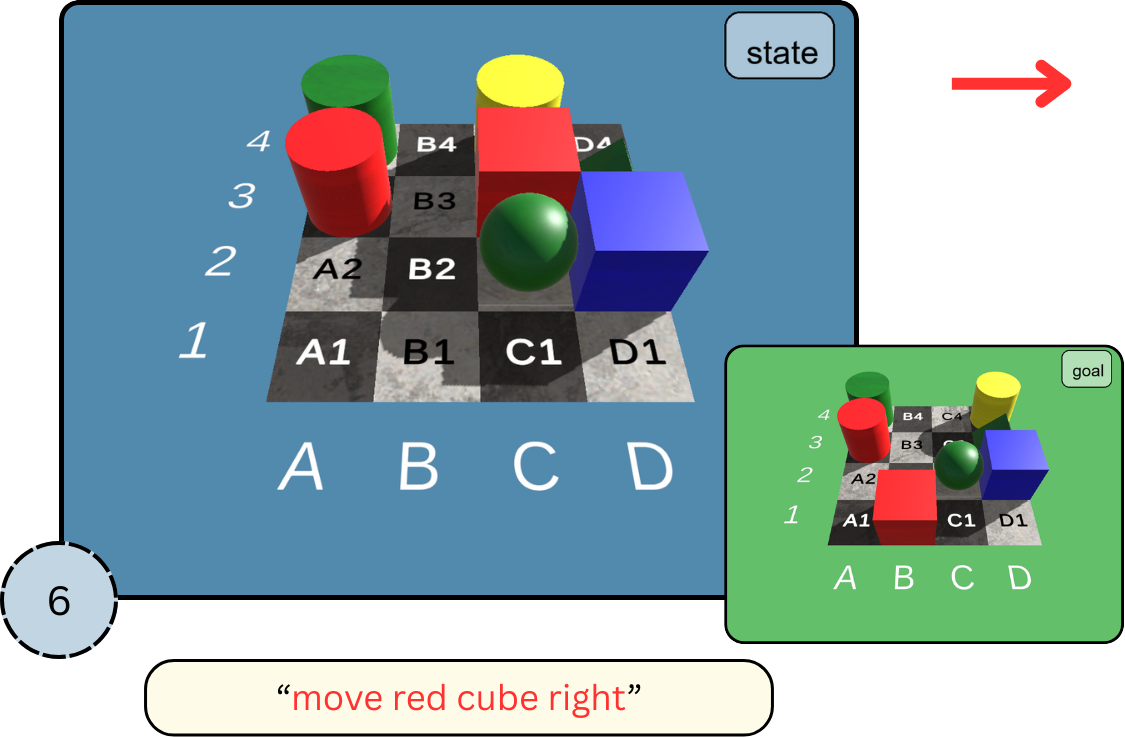}
    \end{subfigure}

    \vskip 0.3in

    \begin{subfigure}{0.32\linewidth}
        \includegraphics[width=\linewidth]{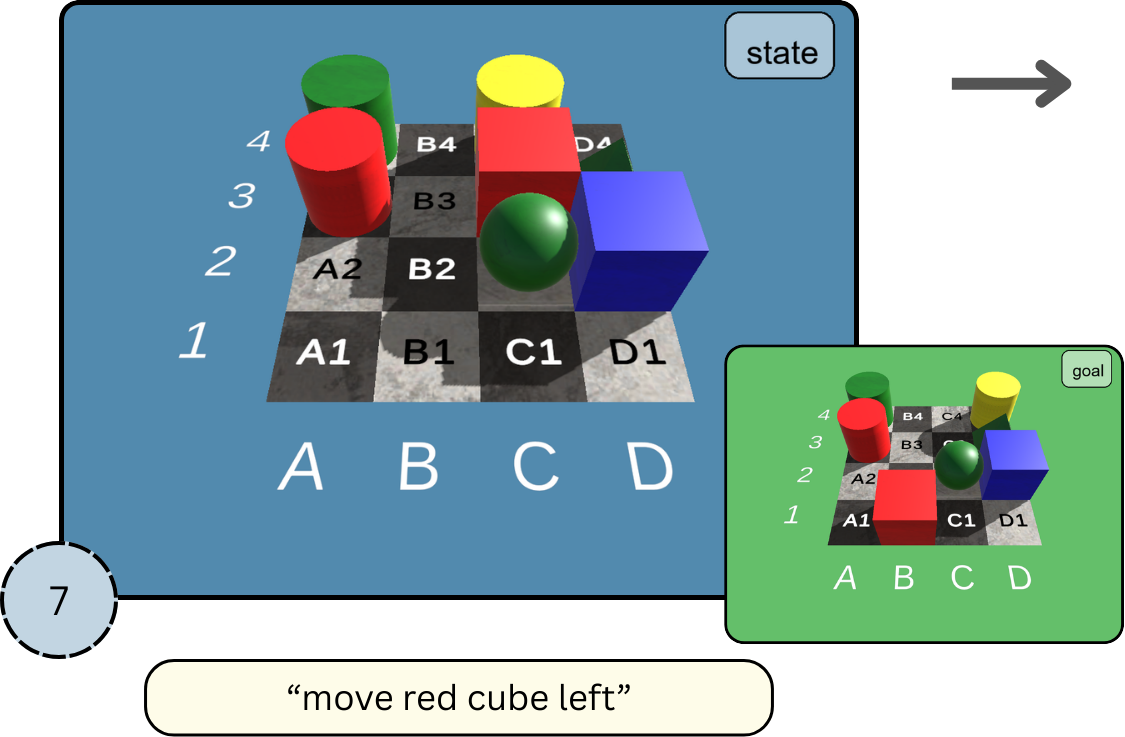}
    \end{subfigure}
    \hfill
    \begin{subfigure}{0.32\linewidth}
        \includegraphics[width=\linewidth]{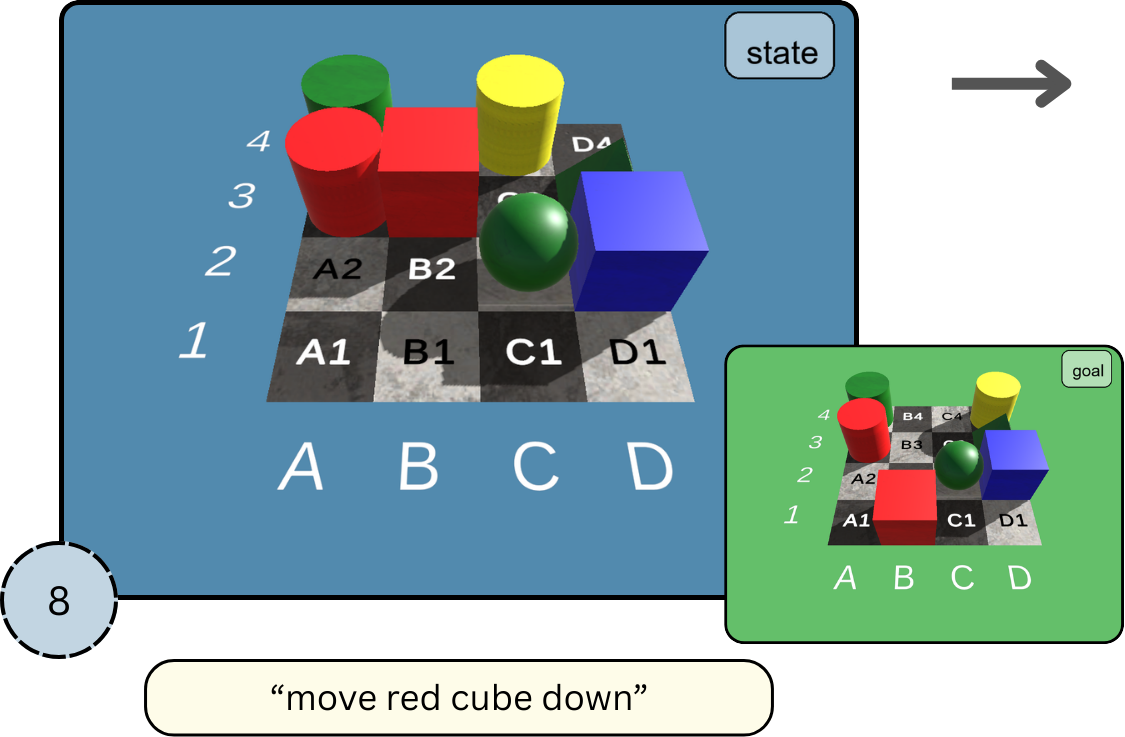}
    \end{subfigure}
    \hfill
    \begin{subfigure}{0.32\linewidth}
        \includegraphics[width=\linewidth]{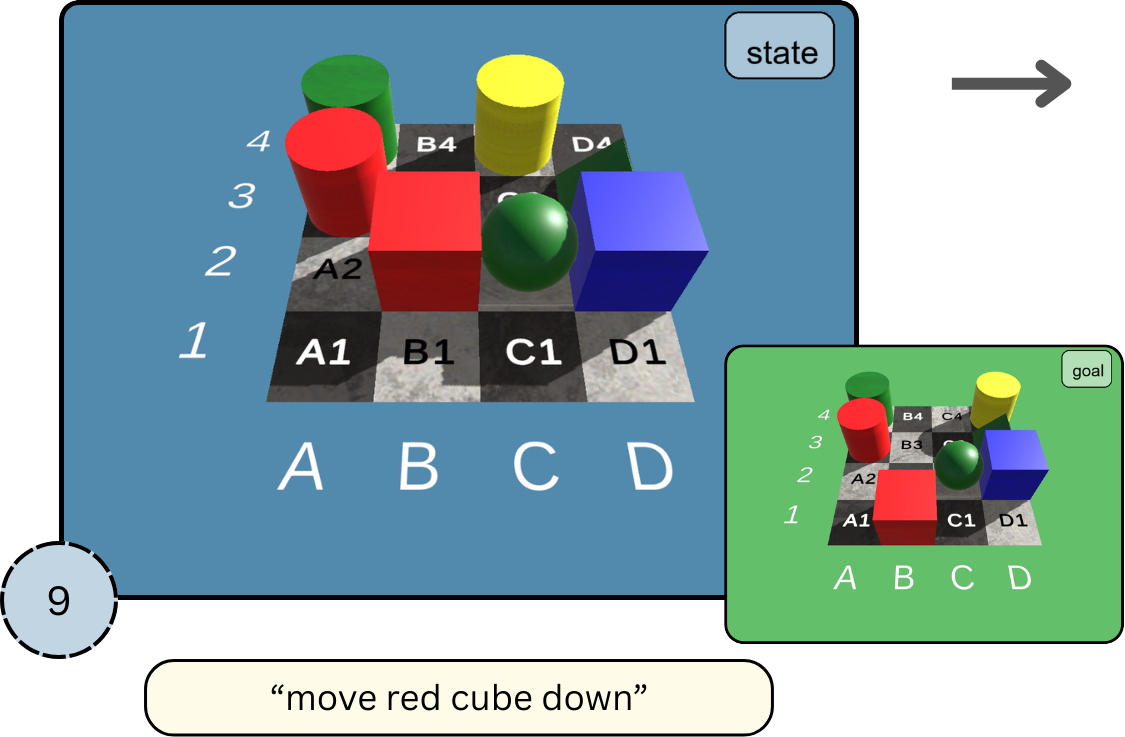}
    \end{subfigure}

    \vskip 0.3in

    \begin{subfigure}{0.32\linewidth}
        \includegraphics[width=\linewidth]{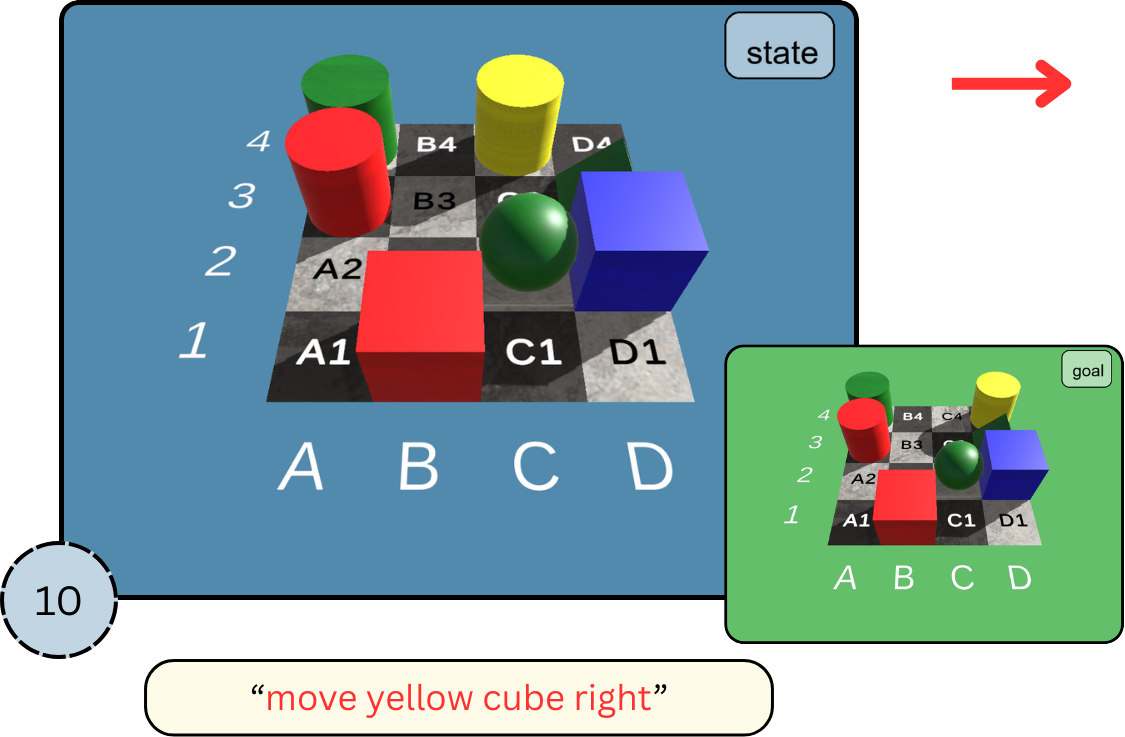}
    \end{subfigure}
    \hfill
    \begin{subfigure}{0.32\linewidth}
        \includegraphics[width=\linewidth]{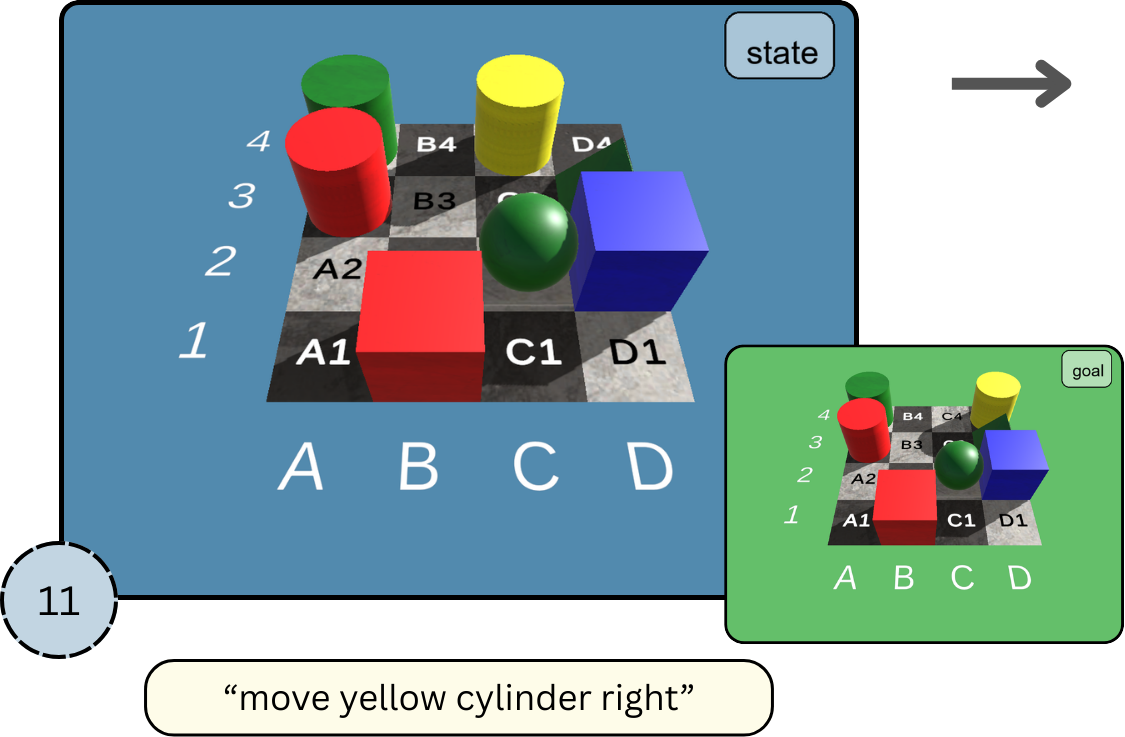}
    \end{subfigure}
    \hfill
    \begin{subfigure}{0.32\linewidth}
        \includegraphics[width=\linewidth]{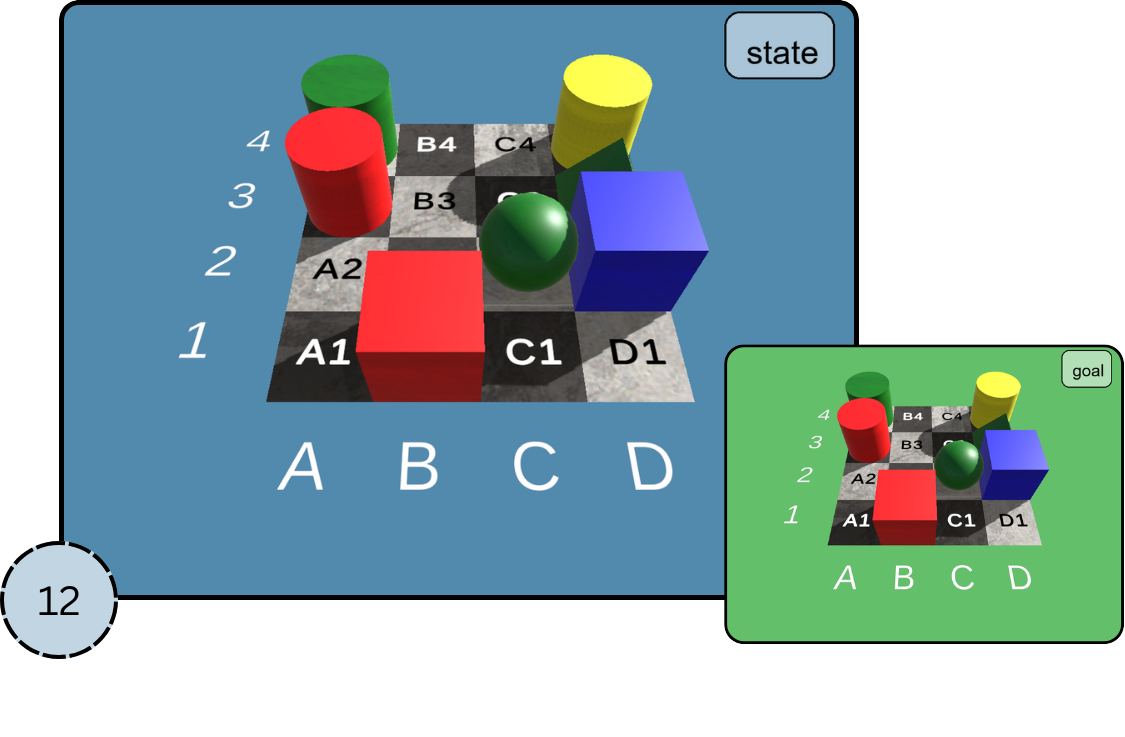}
    \end{subfigure}

    \vskip 0.2in
    \caption{Example of an episode progression for an environment in vision 3D (other modalities progress analogously) with an optimal path length of 9, showing steps 1 to 12 in order, including 3 mistakes (red action text).}
    \label{fig:episode_progression}
\end{figure*}

\newpage
\subsection{Models}
\label{app:models}

\begin{table*}[h]
    \begin{center}
    \resizebox{\textwidth}{!}{%
    \begin{tabular}{l|l|l|l}
        \toprule
        \textbf{Name} & \textbf{LLM} & \textbf{Vision Encoder} & \textbf{Model Size} \\
        \midrule

        \arrayrulecolor{black}\specialrule{0.5pt}{0pt}{0pt}
        \rowcolor{custom_blue}
        \multicolumn{4}{c}{\textbf{Closed Source Models}} \\
        \arrayrulecolor{black}\specialrule{0.5pt}{0pt}{0pt}
        
        Sonnet-3.5 \cite{claudesonnet35} & -- & -- & -- \\
        Gemini-2.0-flash \cite{gemini20flash} & -- & -- & -- \\
        GPT-4o \cite{gpt4o} & -- & -- & -- \\

        \arrayrulecolor{black}\specialrule{0.5pt}{0pt}{0pt}
        \rowcolor{custom_green}
        \multicolumn{4}{c}{\textbf{Open Source Models}} \\
        \arrayrulecolor{black}\specialrule{0.5pt}{0pt}{0pt}
        
        InternVL 2.5 \cite{chen2024expanding} & Qwen 2.5 \cite{qwen2.5} & InternViT \cite{chen2024expanding} & 78.4B \\
        LLaVA OneVision \cite{li2024llava} & Qwen 2 \cite{qwen2} & SigLIP \cite{zhai2023sigmoid} & 73.2B \\
        Qwen 2 VL \cite{Qwen2VL} & Qwen 2 \cite{qwen2} & ViT \cite{Dosovitskiy2020AnII} & 73.4B \\
        \bottomrule
    \end{tabular}%
    }
    \end{center}
    \caption{Overview of evaluated models. -- indicates unavailable information.}
    \label{tab:models}
\end{table*}

\subsection{Sliding Tile Puzzle}
\label{slidingtilepuzzle}

\begin{figure}[h]
    \vskip 0.2in
    \begin{center}
        \centerline{
            \begin{minipage}{0.3\linewidth}
                \centering
                \includegraphics[width=\linewidth]{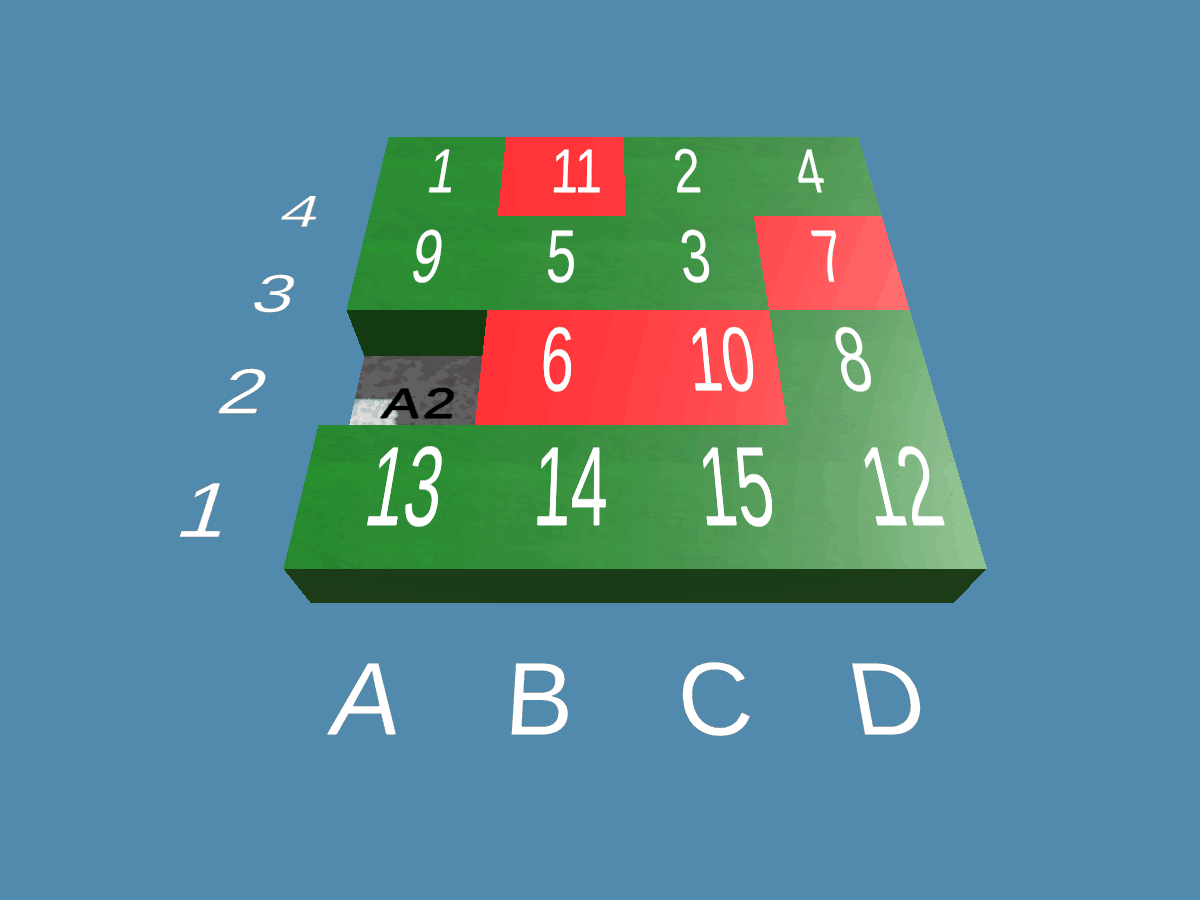}
            \end{minipage}
            \begin{minipage}{0.3\linewidth}
                \centering
                \includegraphics[width=\linewidth]{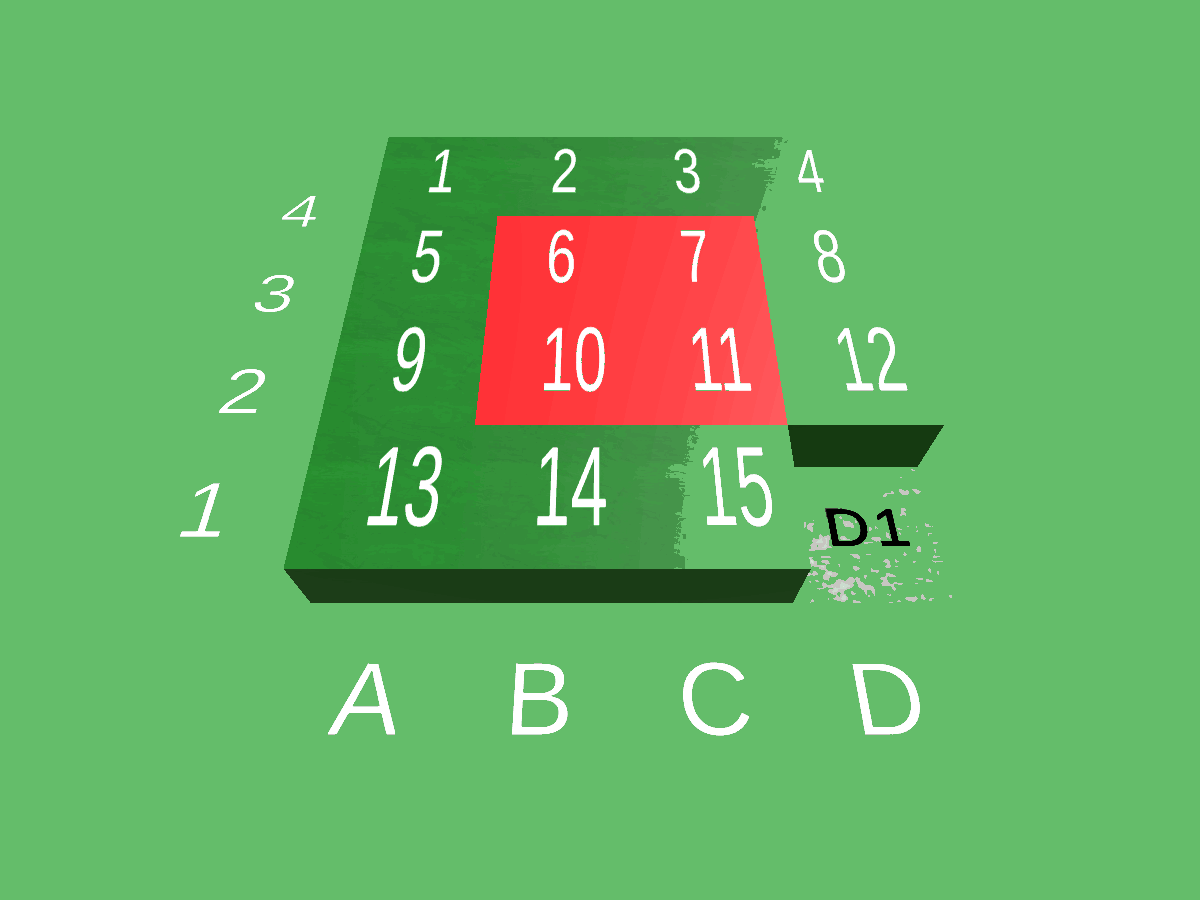}
            \end{minipage}
        }
        \caption{Visualization of a current state and the goal state in a classic 15-tile Sliding Tile Puzzle (STP) on a $4 \times 4$ board, playable by agents within the iVISPAR benchmark.}
        \label{fig:STP_and_geom_vis}
    \end{center}
    \vskip -0.2in
\end{figure}
\begin{flushleft}
The sequential generalized sliding-tile puzzle (SGSTP) is a generalization of the classic 15-Tile Sliding Tile Puzzle (STP), see Figure \ref{fig:STP_and_geom_vis}. In the SGSTP, a set of \( n < m_1 \times m_2 \) tiles, each uniquely labeled \( 1, \dots, n \), are placed on a rectangular grid of size \( m_1 \times m_2 \), denoted by \( G = (V, E) \). The grid has \( m_1 \times m_2 - n \) empty positions that allow tile movement.  

A configuration of tiles is represented as an injective mapping from the set \( \{1, \dots, n\} \) to positions \( V = \{(v_x, v_y) : 1 \leq v_x \leq m_2, 1 \leq v_y \leq m_1 \} \). Each tile must be repositioned from an arbitrary initial configuration \( S = \{s_1, \dots, s_n\} \) to a specified goal configuration \( G = \{g_1, \dots, g_n\} \), such as an ordered row-major layout.  

Let the movement path of tile \( i \), where \( 1 \leq i \leq n \), be expressed as \( p_i : \mathbb{N}_0 \to V \). The puzzle seeks a set of feasible paths \( P = \{p_1, \dots, p_n\} \) that satisfy the following conditions for all \( 1 \leq i, j \leq n \) with \( i \neq j \), and for all time steps \( t \geq 0 \):  

\textbf{Incremental Movement:} \( p_i(t+1) = p_i(t) \text{ or } (p_i(t+1), p_i(t)) \in E \).  
    Tiles move to adjacent, unoccupied positions or stay still.  \\
\textbf{Goal Achievement:} \( p_i(0) = s_i \text{ and } p_i(T) = g_i \text{ for some } T \geq 0 \). 
    Each tile must start at \( s_i \) and reach \( g_i \). \\ 
\textbf{Exclusive Occupancy:} \( p_i(t) \neq p_j(t) \text{ for all } i \neq j \). 
    Two tiles cannot occupy the same position at the same time.  

In this sequential version, tiles move one at a time. Therefore, the head-on collision and corner-following constraints found in the generalized sliding-tile puzzle are omitted, as simultaneous tile movements are not permitted.
\end{flushleft}

\newpage
\subsection{Detailed Results}
\label{sec:detailed_results}
\subsubsection{Performance Results}

\begin{table}[h]
\centering
\setlength{\tabcolsep}{11pt} 
\renewcommand{\arraystretch}{1.1} 

\begin{tabular}{l l cccc}
    \toprule
    \textbf{Model} & \textbf{Metric} & \textbf{Avg} & \textbf{3D} & \textbf{2D} & \textbf{Text} \\
    \midrule

    \arrayrulecolor{black}\specialrule{0.5pt}{0pt}{0pt}
    \rowcolor{custom_blue}
    \multicolumn{6}{c}{\textbf{Closed Source Models}} \\
    \arrayrulecolor{black}\specialrule{0.5pt}{0pt}{0pt}

    \multirow{3}{*}{Sonnet-3.5} 
        & Completed episodes     & 54.56 & 28.67 & 89.67 & 45.33 \\
        & Optimal path deviation & 3.05  & 4.10  & 1.44  & 3.60  \\
        & Board state inference  & 60.00 & 35.38 & 84.62 & --    \\
    \midrule
    \multirow{3}{*}{Gemini-2.0-flash} 
        & Completed episodes     & 27.11 & 12.67 & 47.33 & 21.33 \\
        & Optimal path deviation & 4.87  & 5.25  & 4.09  & 5.26  \\
        & Board state inference  & 54.08 & 28.67 & 79.49 & --    \\
    \midrule
    \multirow{3}{*}{GPT-4o} 
        & Completed episodes     & 17.56 & 9.33  & 37.33 & 6.00  \\
        & Optimal path deviation & 5.30  & 5.45  & 4.15  & 6.30  \\
        & Board state inference  & 41.67 & 19.49 & 63.85 & --    \\
    
    \arrayrulecolor{black}\specialrule{0.5pt}{0pt}{0pt}
    \rowcolor{custom_green}
    \multicolumn{6}{c}{\textbf{Open Source Models}} \\
    \arrayrulecolor{black}\specialrule{0.5pt}{0pt}{0pt}

    \multirow{3}{*}{InternVL2.5-78B} 
        & Completed episodes     & 10.16 & 1.67  & 9.42  & 19.33 \\
        & Optimal path deviation & 5.98  & 6.39  & 5.86  & 5.69  \\
        & Board state inference  & 34.95 & 16.51 & 53.38 & --    \\
    \midrule
    \multirow{3}{*}{LLaVA-OneVision-72B} 
        & Completed episodes     & 8.22  & 0.67  & 1.33  & 22.67 \\
        & Optimal path deviation & 6.35  & 6.75  & 6.81  & 5.50  \\
        & Board state inference  & 26.36 & 14.72 & 38.00 & --    \\
    \midrule
    \multirow{3}{*}{Qwen2-72B} 
        & Completed episodes     & 5.89  & 0.67  & 1.67  & 15.33 \\
        & Optimal path deviation & 6.37  & 6.66  & 6.54  & 5.90  \\
        & Board state inference  & 41.54 & 18.77 & 64.31  & --    \\
    
    \arrayrulecolor{black}\specialrule{0.5pt}{0pt}{0pt}
    \rowcolor{custom_gray}
    \multicolumn{6}{c}{\textbf{Aggregate Averages}} \\
    \arrayrulecolor{black}\specialrule{0.5pt}{0pt}{0pt}
    \multirow{3}{*}{\textit{Average}} 
        & Completed episodes     & 20.59 & 7.04  & 26.68 & 21.83 \\
        & Optimal path deviation & 5.32  & 5.76  & 4.41  & 5.32  \\
        & Board state inference  & 43.10 & 22.26 & 63.94 & --    \\

    \bottomrule
\end{tabular}%

\caption{Evaluation of models across three modalities. Each row shows average episode completion rate (\%), mean deviation from the optimal path (see Section \ref{sec:eval}), and board state inference accuracy (\%).}
\label{tab:VLM_main_statistics_restructured}
\end{table}

\newpage

\subsubsection{Error Counts for the Geom Puzzle}

\begin{table}[h]
\centering

\setlength{\tabcolsep}{18pt} 
\renewcommand{\arraystretch}{.85} 

\begin{tabular}{l l cccc}
    \toprule
    \textbf{Model} & \textbf{Metric} & \textbf{Avg} & \textbf{3D} & \textbf{2D} & \textbf{Text} \\
    \midrule

    \arrayrulecolor{black}\specialrule{0.5pt}{0pt}{0pt}
    \rowcolor{custom_blue}
    \multicolumn{6}{c}{\textbf{Closed Source Models}} \\
    \arrayrulecolor{black}\specialrule{0.5pt}{0pt}{0pt}

    \multirow{5}{*}{Sonnet-3.5}
        & EM                    & 6.31 & 6.51 & 6.20 & 6.21 \\
        & IM                    & 1.86 & 3.34 & 0.21 & 2.03 \\
        & OD                    & 3.60 & 4.77 & 2.29 & 3.75 \\
        & OB                    & 1.59 & 1.95 & 0.04 & 2.79 \\
        & IC                    & 0.02 & 0.07 & 0.00 & 0.00 \\
    \midrule
    \multirow{5}{*}{Gemini-2.0-flash}
        & EM                    & 5.68 & 5.80 & 6.34 & 4.91 \\
        & IM                    & 2.95 & 3.87 & 2.35 & 2.63 \\
        & OD                    & 6.14 & 6.83 & 5.51 & 6.08 \\
        & OB                    & 2.56 & 2.25 & 1.11 & 4.33 \\
        & IC                    & 0.01 & 0.01 & 0.00 & 0.03 \\
    \midrule
    \multirow{5}{*}{GPT-4o}
        & EM                    & 4.65 & 5.50 & 5.95 & 2.51 \\
        & IM                    & 2.86 & 4.03 & 2.36 & 2.19 \\
        & OD                    & 6.53 & 6.36 & 5.51 & 7.71 \\
        & OB                    & 3.81 & 2.69 & 1.85 & 6.90 \\
        & IC                    & 0.26 & 0.24 & 0.52 & 0.03 \\

    \arrayrulecolor{black}\specialrule{0.5pt}{0pt}{0pt}
    \rowcolor{custom_green}
    \multicolumn{6}{c}{\textbf{Open Source Models}} \\
    \arrayrulecolor{black}\specialrule{0.5pt}{0pt}{0pt}

    \multirow{5}{*}{InternVL2.5-78B}
        & EM                    & 5.00 & 4.94 & 5.74 & 4.39 \\
        & IM                    & 4.24 & 5.39 & 4.80 & 2.59 \\
        & OD                    & 5.90 & 6.06 & 5.70 & 5.92 \\
        & OB                    & 3.38 & 3.16 & 2.52 & 4.38 \\
        & IC                    & 0.59 & 0.21 & 0.29 & 1.26 \\
    \midrule
    \multirow{5}{*}{LLaVA-OneVision-72B}
        & EM                    & 3.95 & 3.41 & 3.22 & 5.23 \\
        & IM                    & 4.12 & 4.55 & 4.42 & 3.40 \\
        & OD                    & 4.89 & 4.58 & 4.74 & 5.36 \\
        & OB                    & 4.17 & 4.62 & 4.46 & 3.44 \\
        & IC                    & 1.38 & 1.90 & 2.19 & 0.07 \\
    \midrule
    \multirow{5}{*}{Qwen2-72B}
        & EM                    & 4.07 & 3.88 & 3.96 & 4.85 \\
        & IM                    & 4.61 & 4.81 & 4.67 & 3.89 \\
        & OD                    & 5.39 & 5.55 & 5.21 & 5.25 \\
        & OB                    & 3.72 & 4.05 & 3.17 & 3.83 \\
        & IC                    & 0.10 & 0.07 & 0.06 & 0.26 \\

    \arrayrulecolor{black}\specialrule{0.5pt}{0pt}{0pt}
    \rowcolor{custom_gray}
    \multicolumn{6}{c}{\textbf{Aggregate Averages}} \\
    \arrayrulecolor{black}\specialrule{0.5pt}{0pt}{0pt}
    \multirow{5}{*}{\textit{Average}}
        & EM                    & 4.82 & 4.72 & 5.04 & 4.68 \\
        & IM                    & 3.61 & 4.45 & 3.34 & 2.79 \\
        & OD                    & 5.40 & 5.66 & 4.87 & 5.68 \\
        & OB                    & 3.28 & 3.35 & 2.33 & 4.28 \\
        & IC                    & 0.35 & 0.33 & 0.45 & 0.28 \\

    \bottomrule
\end{tabular}%
\caption{Evaluation of models across three modalities. Each row shows average steps per episode that were effective moves (EM), ineffective moves (IM), occupied destination moves (OD), out of bounds moves (OB) and illegal commands (IC).}
\label{tab:VLM_main_experiment_error_counts}
\end{table}

\newpage

\subsubsection{Error Counts for the Auxiliary Task}
\begin{table}[h]
\centering
\setlength{\tabcolsep}{18pt} 
\renewcommand{\arraystretch}{.9} 
\begin{tabular}{l l cccc}
    \toprule
    \textbf{Model} & \textbf{Metric} & \textbf{Avg} & \textbf{3D} & \textbf{2D} & \textbf{Text} \\
    \midrule

    \arrayrulecolor{black}\specialrule{0.5pt}{0pt}{0pt}
    \rowcolor{custom_blue}
    \multicolumn{6}{c}{\textbf{Closed Source Models}} \\
    \arrayrulecolor{black}\specialrule{0.5pt}{0pt}{0pt}

    \multirow{7}{*}{Sonnet-3.5}
        & Correct            & 3.90 & 2.30 & 5.50 & -- \\
        & Missed             & 1.42 & 1.84 & 1.00 & -- \\
        & Hallucinated       & 0.00 & 0.00 & 0.00 & -- \\
        & Coord Errors       & 1.08 & 2.16 & 0.00 & -- \\
        & Color Errors       & 0.38 & 0.76 & 0.00 & -- \\
        & Shape Errors       & 0.37 & 0.74 & 0.00 & -- \\
        & Format Errors      & 0.00 & 0.00 & 0.00 & -- \\
    \midrule
    \multirow{7}{*}{Gemini-2.0-flash}
        & Correct            & 3.52 & 1.86 & 5.17 & -- \\
        & Missed             & 0.91 & 1.02 & 0.80 & -- \\
        & Hallucinated       & 0.14 & 0.13 & 0.14 & -- \\
        & Coord Errors       & 1.98 & 3.48 & 0.48 & -- \\
        & Color Errors       & 0.66 & 1.14 & 0.18 & -- \\
        & Shape Errors       & 0.65 & 1.14 & 0.16 & -- \\
        & Format Errors      & 0.05 & 0.00 & 0.09 & -- \\
    \midrule
    \multirow{7}{*}{GPT-4o}
        & Correct            & 2.71 & 1.27 & 4.15 & -- \\
        & Missed             & 1.31 & 1.67 & 0.95 & -- \\
        & Hallucinated       & 0.03 & 0.01 & 0.04 & -- \\
        & Coord Errors       & 2.34 & 3.33 & 1.35 & -- \\
        & Color Errors       & 0.77 & 1.18 & 0.35 & -- \\
        & Shape Errors       & 0.75 & 1.18 & 0.32 & -- \\
        & Format Errors      & 0.02 & 0.04 & 0.00 & -- \\

    \arrayrulecolor{black}\specialrule{0.5pt}{0pt}{0pt}
    \rowcolor{custom_gray}
    \multicolumn{6}{c}{\textbf{Aggregate Averages}} \\
    \arrayrulecolor{black}\specialrule{0.5pt}{0pt}{0pt}

    \multirow{7}{*}{\textit{Average}}
        & Correct            & 3.37 & 1.81 & 4.94 & -- \\
        & Missed             & 1.21 & 1.51 & 0.92 & -- \\
        & Hallucinated       & 0.06 & 0.05 & 0.06 & -- \\
        & Coord Errors       & 1.80 & 2.99 & 0.61 & -- \\
        & Color Errors       & 0.60 & 1.03 & 0.18 & -- \\
        & Shape Errors       & 0.59 & 1.02 & 0.16 & -- \\
        & Format Errors      & 0.02 & 0.01 & 0.03 & -- \\

    \bottomrule
\end{tabular}%

\caption{Error analysis for the auxiliary position inference task across vision modalities (closed source models).}
\label{tab:auxiliary_position_inference_errors_closed}
\end{table}

\newpage

\begin{table}[h]
\centering
\setlength{\tabcolsep}{16pt} 
\renewcommand{\arraystretch}{.9} 
\begin{tabular}{l l cccc}
    \toprule
    \textbf{Model} & \textbf{Metric} & \textbf{Avg} & \textbf{3D} & \textbf{2D} & \textbf{Text} \\
    \midrule

    \arrayrulecolor{black}\specialrule{0.5pt}{0pt}{0pt}
    \rowcolor{custom_green}
    \multicolumn{6}{c}{\textbf{Open Source Models}} \\
    \arrayrulecolor{black}\specialrule{0.5pt}{0pt}{0pt}

    \multirow{7}{*}{InternVL2.5-78B}
        & Correct            & 2.27 & 1.07 & 3.47 & -- \\
        & Missed             & 0.89 & 1.00 & 0.77 & -- \\
        & Hallucinated       & 0.03 & 0.04 & 0.01 & -- \\
        & Coord Errors       & 1.62 & 2.92 & 0.32 & -- \\
        & Color Errors       & 0.59 & 1.11 & 0.08 & -- \\
        & Shape Errors       & 0.58 & 1.08 & 0.08 & -- \\
        & Format Errors      & 1.63 & 1.30 & 1.97 & -- \\
    \midrule
    \multirow{7}{*}{LLaVA-OneVision-72B}
        & Correct            & 1.71 & 0.96 & 2.47 & -- \\
        & Missed             & 1.02 & 1.18 & 0.86 & -- \\
        & Hallucinated       & 0.34 & 0.31 & 0.37 & -- \\
        & Coord Errors       & 3.30 & 3.95 & 2.65 & -- \\
        & Color Errors       & 1.28 & 1.58 & 0.97 & -- \\
        & Shape Errors       & 1.23 & 1.57 & 0.90 & -- \\
        & Format Errors      & 0.37 & 0.09 & 0.65 & -- \\
    \midrule
    \multirow{7}{*}{Qwen2-72B}
        & Correct            & 2.70 & 1.22 & 4.18 & -- \\
        & Missed             & 0.97 & 1.08 & 0.85 & -- \\
        & Hallucinated       & 0.58 & 0.81 & 0.36 & -- \\
        & Coord Errors       & 2.52 & 3.80 & 1.24 & -- \\
        & Color Errors       & 0.93 & 1.42 & 0.43 & -- \\
        & Shape Errors       & 1.12 & 1.67 & 0.58 & -- \\
        & Format Errors      & 0.22 & 0.06 & 0.38 & -- \\

    \arrayrulecolor{black}\specialrule{0.5pt}{0pt}{0pt}
    \rowcolor{custom_gray}
    \multicolumn{6}{c}{\textbf{Aggregate Averages}} \\
    \arrayrulecolor{black}\specialrule{0.5pt}{0pt}{0pt}

    \multirow{7}{*}{\textit{Average}}
        & Correct            & 2.23 & 1.08 & 3.37 & -- \\
        & Missed             & 0.96 & 1.09 & 0.83 & -- \\
        & Hallucinated       & 0.32 & 0.39 & 0.25 & -- \\
        & Coord Errors       & 2.48 & 3.56 & 1.40 & -- \\
        & Color Errors       & 0.93 & 1.37 & 0.49 & -- \\
        & Shape Errors       & 0.98 & 1.44 & 0.52 & -- \\
        & Format Errors      & 0.74 & 0.48 & 1.00 & -- \\

    \bottomrule
\end{tabular}%

\caption{Error analysis for the auxiliary position inference task across vision modalities (open source models).}
\label{tab:auxiliary_position_inference_errors_open}
\end{table}


\newpage

\subsection{Supplementary Graphs}
\label{sec:sup_graphs}

\begin{figure*}[ht]
    \vskip 0.2in
    \begin{center}
        \centerline{\includegraphics[width=\linewidth]{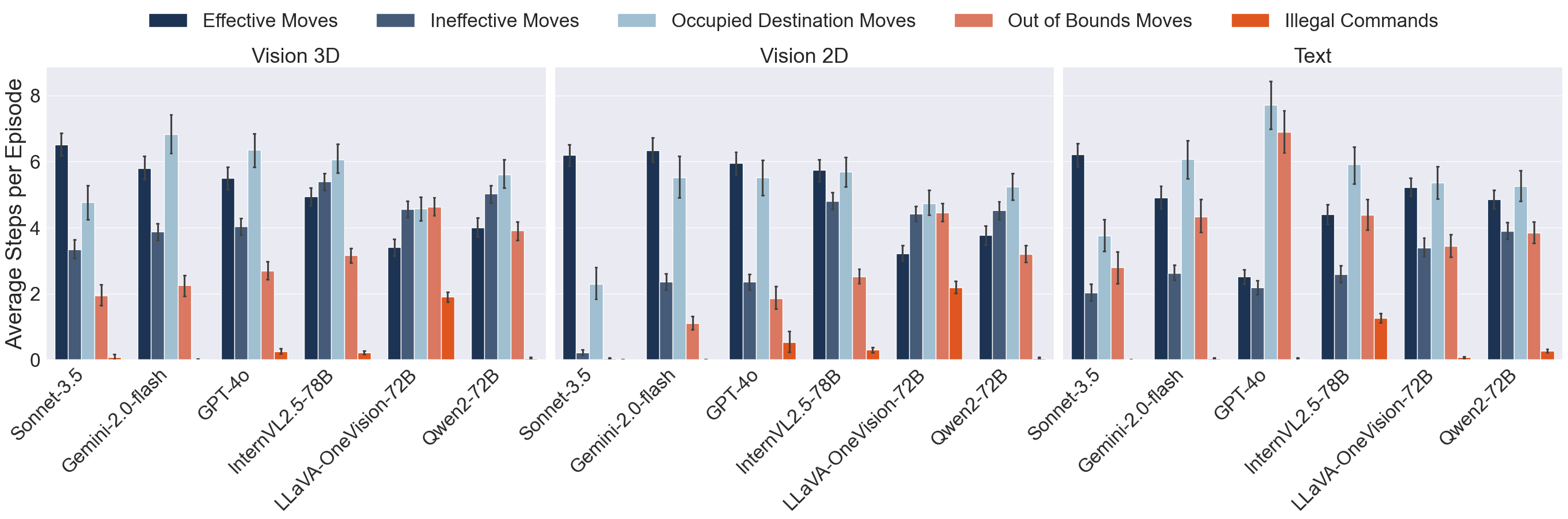}}
        \caption{VLMs' average action counts per episode by category for each modality. Number of actions per episode is capped at 20. Effective / ineffective actions respectively decrease / increase the path length to the goal state. Occupied destination and out-of-bounds are invalid moves, while illegal commands break the instructed action format, all of which leave the board state unchanged.}
        \label{fig:barplot_action_counts}
    \end{center}
    \vskip -0.2in
\end{figure*}

\begin{figure*}[h]
    \vskip 0.2in
    \begin{center}
        \centerline{\includegraphics[width=\linewidth]{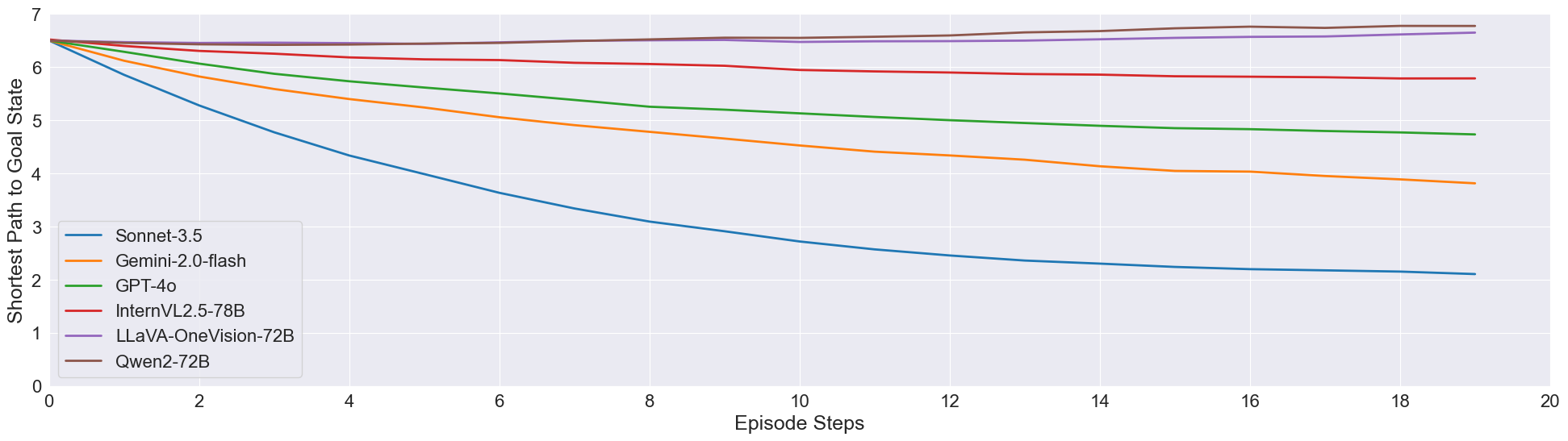}}
        \caption{VLMs' average shortest path to the goal state across all modalities. Number of actions per episode is capped at 20.}
        \label{fig:lineplot_episode_progress}
    \end{center}
    \vskip -0.2in
\end{figure*}



\newpage
\twocolumn
\subsection{Additional Agent Interaction Data}

\subsubsection{Systematic Formatting Errors}
\label{sec:invalid_commands}

Unless noted otherwise, the numeral in parentheses after a model name is the \emph{count of formatting errors} for that category. Notably, Sonnet-3.5 is not listed since it did not make any format errors, explaining its high benchmarking score.

\paragraph{(E1) Empty-cell mentions (\textbf{N = 280})}%
\mbox{}\\
The most common violation is the explicit listing of empty grid cells, even though instructions forbid any mention of empties.  Surface forms vary widely, even within a single model:

\begin{colorbox_no_header}
a4 empty
\end{colorbox_no_header}

\begin{colorbox_no_header}
c3 blank
\end{colorbox_no_header}

\begin{colorbox_no_header}
b1 no object
\end{colorbox_no_header}




\begin{colorbox_no_header}
a1 none none
\end{colorbox_no_header}

\noindent Gemini-2.0-flash (24), InternVL-2.5-78B (21), LLaVA-OneVision-72B (105), Qwen2-72B (130).

\paragraph{(E2) Missing attributes (\textbf{N = 88})}%
\mbox{}\\
Some lines list an object but drop one of its required attributes (colour or shape):

\begin{colorbox_no_header}
c1 none pyramid
\end{colorbox_no_header}

\begin{colorbox_no_header}
b2 sphere
\end{colorbox_no_header}

\noindent
Gemini-2.0-flash (2), LLaVA-OneVision-72B (86).

\paragraph{(E3) Illegal attributes (\textbf{N = 21})}%
\mbox{}\\
Entries introduce colours or shapes outside the predefined vocabulary, or mis-name legitimate ones:


\begin{colorbox_no_header}
b2 black cone
\end{colorbox_no_header}

\begin{colorbox_no_header}
b3 red block
\end{colorbox_no_header}


\noindent
Gemini-2.0-flash (2), GPT-4o (9), LLaVA-OneVision-72B (9), Qwen2-72B (1).

\paragraph{(E4) Attribute order inversion (\textbf{N = 21})}%
\mbox{}\\
Colour–shape order is sometimes reversed:

\begin{colorbox_no_header}
b4 cylinder yellow
\end{colorbox_no_header}

\noindent LLaVA-OneVision-72B (21)

\paragraph{(E5) Multiple attributes per object (\textbf{N = 3})}%
\mbox{}\\
A single coordinate is assigned more than one colour or a compound colour phrase:

\begin{colorbox_no_header}
d4 blue and red cube
\end{colorbox_no_header}

\begin{colorbox_no_header}
d1 multicoloured cube
\end{colorbox_no_header}

\begin{colorbox_no_header}
a2 a3 green pyramid
\end{colorbox_no_header}

\noindent
GPT-4o (1), InternVL-2.5-78B (1), Qwen2-72B (1).

\paragraph{(E6) Out-of-bounds coordinates (\textbf{N = 3})}%
\mbox{}\\
Objects occasionally appear outside the 4\,$\times$\,4 grid, e.g.\ likely due to OCR misreads:

\begin{colorbox_no_header}
f4 yellow pyramid
\end{colorbox_no_header}

\begin{colorbox_no_header}
c5 green pyramid
\end{colorbox_no_header}

\noindent
Gemini-2.0-flash (1), InternVL-2.5-78B (1), LLaVA-OneVision-72B (1).

\paragraph{(E7) Non-list answers (\textbf{N = 47})}%
\mbox{}\\
A few generations present free-form summaries or spurious URLs instead of the required flat list:

\begin{colorbox_no_header}
d2 blue cylinder summary a4 yellow cylinder b4 blue sphere \dots
\end{colorbox_no_header}

\begin{colorbox_no_header}
d3 green sphere https user images githubusercontent com /.../f45e3211ecb93a6b429446.png
\end{colorbox_no_header}

\begin{colorbox_no_header}
obj in objectsitems if obj none output
\end{colorbox_no_header}

\begin{colorbox_no_header}
yellow objectshape must be exactly one of cube
\end{colorbox_no_header}

\begin{colorbox_no_header}
we will create the solution
\end{colorbox_no_header}

\noindent
InternVL-2.5-78B (47).

\paragraph{(E8) Solution not found (\textbf{N = 147})}%
\mbox{}\\
Prompts that did not begin their solution with "Solution: " are not picked up by the solution parser.

\noindent  GPT-4o (1), InternVL-2.5-78B (145), Qwen2-72B (1).

\newpage
\subsubsection{Chain of Thoughts}
\label{sec:chain_of_thoughts}

\begin{colorbox_header}{Sonnet-3.5 CoT Example}{custom_green}
Looking at the current state and goal state:

\begin{enumerate}[nosep,leftmargin=*]
\item The blue cube is correctly positioned in A2 in both states. 
\item The yellow pyramid needs to move from D1 to C1. 
\item The red cylinder is in D2 and needs to be in D3. 
\end{enumerate}

The yellow pyramid needs to be moved to make space for other movements. Looking at the past states we tried moving the yellow pyramid right and the red cylinder up. Now we need to move the yellow pyramid to its goal position. I can see that the yellow pyramid should move left to reach position C1 as shown in the goal state.
\end{colorbox_header}

\begin{colorbox_header}{Qwen2-72B CoT Example}{custom_green}
After analyzing the images, it is clear that moving the red sphere down was an invalid action because it resulted in no change to the board state. The red sphere is still in the same position, and the green cylinder has not moved.

To progress, I need to find a valid action that will move one of the objects to match the goal state.

Upon further inspection, I notice that the red sphere needs to be moved to the top-right corner of the board to match the goal state. However, the current state shows the red sphere in the bottom-right corner, which means I need to move it up to get it closer to the goal position.
\end{colorbox_header}

\begin{colorbox_header}{GPT-4o CoT Example}{custom_green}
To reach the goal state I need to compare the current active state with the goal state.

In the current active state the green sphere is on top blocking movements of objects below it. The yellow sphere is on the bottom row B1.

In the goal state the green sphere is two steps down in column B. The red cylinder is in the bottom left, two steps down. The yellow sphere is in position B2, two steps up.

Analyzing the images I see that the yellow spheres path is blocked by the green cylinder I’ll first move another object to clear the way. Moving the blue cube rightward will help clear the blocked pathway for the green cylinder.
\end{colorbox_header}

\begin{colorbox_header}{InternVL-2.5-78B CoT Example}{custom_green}
to reach the goal state, let's analyze the current board and the objects that need to be moved:

\textbf{current state (active)}
\begin{itemize}[nosep,leftmargin=*]
\item the red cylinder is on tile b4.
\item the yellow pyramid is on tile d4.
\item the blue cylinder is on tile c3.
\item the yellow cube is on tile d2.
\end{itemize}

\textbf{goal state (goal)}
\begin{itemize}[nosep,leftmargin=*]
\item the red cylinder needs to be on tile a4.
\item the blue cylinder needs to be on tile b4.
\item the yellow cube needs to be on tile d3.
\item the yellow pyramid needs to be on tile c4.
\end{itemize}

\textbf{step-by-step plan}

1. move the red cylinder to a4:\\
\quad - action: move the red cylinder left.\\[4pt]
2. move the yellow pyramid to c4:\\
\quad - action: move the yellow pyramid down.\\[4pt]
3. move the yellow cube to d3:\\
\quad - action: move the yellow cube down.\\[4pt]
4. move the blue cylinder to b4:\\
\quad - action: move the blue cylinder up.

\textbf{first action}\\
the first step is to move the red cylinder from tile b4 to tile a4.
\end{colorbox_header}

\begin{colorbox_header}{Gemini-2.0-flash CoT Example}{custom_green}

The green sphere needs to move from B2 to C3. It is currently at B2, moving up would move the green sphere to B3 where it is blocked by the red pyramid. 

I need to move the green sphere to the right, then up. Its already been moved right once, so my next action is to move the green sphere up.

\end{colorbox_header}

\begin{colorbox_header}{LLaVA-OneVision-72B CoT Example}{custom_green}
The blue cube on \texttt{D2} needs to move to the position of the red cylinder on \texttt{D1}.  

Since the red cylinder cannot be displaced, the blue cube must be moved elsewhere.  

Moving the blue cube left to \texttt{C2} does not change the board state because another blue cube already occupies that square.  

Moving it up to \texttt{D1} would push the red cylinder out of bounds, which is not allowed. 

Therefore, the blue cube on \texttt{D2} should move left to \texttt{D3}.
\end{colorbox_header}

\onecolumn




\newpage
\subsection{Algorithms}
\label{sec:algorithms}

\subsubsection{Optimal Agent}
\label{sec:optimal_agent}
\begin{algorithm}
\caption{A* Algorithm for Pathfinding}
\label{alg:astar}
\begin{algorithmic}[1]
\Require start, goal
\Ensure Path from start to goal or failure

\State openSet $\gets$ \{start\}
\State cameFrom $\gets$ empty map
\State gScore[start] $\gets$ 0
\State fScore[start] $\gets$ heuristic(start, goal)

\While{openSet not empty}
    \State current $\gets$ node in openSet with lowest fScore
    \If{current = goal}
        \State \Return ReconstructPath(cameFrom, current)
    \EndIf
    \State Remove current from openSet

    \For{each neighbor of current}
        \State tentativeGScore $\gets$ gScore[current] + d(current, neighbor)
        \If{tentativeGScore < gScore[neighbor] \textbf{or} neighbor not in gScore}
            \State cameFrom[neighbor] $\gets$ current
            \State gScore[neighbor] $\gets$ tentativeGScore
            \State fScore[neighbor] $\gets$ gScore[neighbor] + heuristic(neighbor, goal)
            \If{neighbor not in openSet}
                \State Add neighbor to openSet
            \EndIf
        \EndIf
    \EndFor
\EndWhile

\State \Return failure
\end{algorithmic}
\end{algorithm}

\subsubsection{Random Agent}
\label{sec:rand_agent}
\begin{algorithm}
\caption{Generate Random Valid Path for Sliding Tile Puzzle}
\label{alg:random_agent}
\begin{algorithmic}[1]
\Require $n$ (board size), initial\_state, max\_steps
\Ensure Path from initial to final state

\State path $\gets$ [initial\_state]
\State current\_state $\gets$ initial\_state

\For{step = 1 to max\_steps}
    \State neighbors $\gets$ get\_neighbors(current\_state, $n$)
    \State current\_state $\gets$ random choice from neighbors
    \State Append current\_state to path
\EndFor

\Return path
\end{algorithmic}
\end{algorithm}


\begin{thebibliography}{57}
\providecommand{\natexlab}[1]{#1}

\bibitem[{Abdou et~al.(2021)Abdou, Kulmizev, Hershcovich, Frank, Pavlick, and Søgaard}]{abdou_can_2021}
Mostafa Abdou, Artur Kulmizev, Daniel Hershcovich, Stella Frank, Ellie Pavlick, and Anders Søgaard. 2021.
\newblock \href {https://arxiv.org/abs/2109.06129v2} {Can language models encode perceptual structure without grounding? a case study in color}.

\bibitem[{Aghzal et~al.(2024)Aghzal, Plaku, and Yao}]{aghzal_can_2024}
Mohamed Aghzal, Erion Plaku, and Ziyu Yao. 2024.
\newblock \href {https://doi.org/10.48550/arXiv.2310.03249} {Can large language models be good path planners? a benchmark and investigation on spatial-temporal reasoning}.
\newblock \emph{Preprint}, arxiv:2310.03249 [cs].

\bibitem[{Bohnet et~al.(2024)Bohnet, Nova, Parisi, Swersky, Goshvadi, Dai, Schuurmans, Fiedel, and Sedghi}]{bohnet_exploring_2024}
Bernd Bohnet, Azade Nova, Aaron~T. Parisi, Kevin Swersky, Katayoon Goshvadi, Hanjun Dai, Dale Schuurmans, Noah Fiedel, and Hanie Sedghi. 2024.
\newblock \href {https://doi.org/10.48550/arXiv.2406.13094} {Exploring and benchmarking the planning capabilities of large language models}.
\newblock \emph{Preprint}, arxiv:2406.13094 [cs].

\bibitem[{Bordes et~al.(2024)Bordes, Pang, Ajay, Li, Bardes, Petryk, Ma{\~{n}}as, Lin, Mahmoud, Jayaraman, Ibrahim, Hall, Xiong, Lebensold, Ross, Jayakumar, Guo, Bouchacourt, Al{-}Tahan, Padthe, Sharma, Xu, Tan, Richards, Lavoie, Astolfi, Hemmat, Chen, Tirumala, Assouel, Moayeri, Talattof, Chaudhuri, Liu, Chen, Garrido, Ullrich, Agrawal, Saenko, Celikyilmaz, and Chandra}]{DBLP:journals/corr/abs-2405-17247}
Florian Bordes, Richard~Yuanzhe Pang, Anurag Ajay, Alexander~C. Li, Adrien Bardes, Suzanne Petryk, Oscar Ma{\~{n}}as, Zhiqiu Lin, Anas Mahmoud, Bargav Jayaraman, Mark Ibrahim, Melissa Hall, Yunyang Xiong, Jonathan Lebensold, Candace Ross, Srihari Jayakumar, Chuan Guo, Diane Bouchacourt, Haider Al{-}Tahan, and 22 others. 2024.
\newblock \href {https://doi.org/10.48550/ARXIV.2405.17247} {An introduction to vision-language modeling}.
\newblock \emph{CoRR}, abs/2405.17247.

\bibitem[{Campbell et~al.(2024)Campbell, Rane, Giallanza, Sabbata, Ghods, Joshi, Ku, Frankland, Griffiths, Cohen, and Webb}]{DBLP:journals/corr/abs-2411-00238}
Declan Campbell, Sunayana Rane, Tyler Giallanza, Nicol{\`{o}}~De Sabbata, Kia Ghods, Amogh Joshi, Alexander Ku, Steven~M. Frankland, Thomas~L. Griffiths, Jonathan~D. Cohen, and Taylor~W. Webb. 2024.
\newblock \href {https://doi.org/10.48550/ARXIV.2411.00238} {Understanding the limits of vision language models through the lens of the binding problem}.
\newblock \emph{CoRR}, abs/2411.00238.

\bibitem[{Campbell et~al.(2025)Campbell, Rane, Giallanza, Sabbata, Ghods, Joshi, Ku, Frankland, Griffiths, Cohen, and Webb}]{campbell_understanding_2025}
Declan Campbell, Sunayana Rane, Tyler Giallanza, Nicolò~De Sabbata, Kia Ghods, Amogh Joshi, Alexander Ku, Steven~M. Frankland, Thomas~L. Griffiths, Jonathan~D. Cohen, and Taylor~W. Webb. 2025.
\newblock \href {https://doi.org/10.48550/arXiv.2411.00238} {Understanding the limits of vision language models through the lens of the binding problem}.
\newblock \emph{Preprint}, arxiv:2411.00238 [cs].

\bibitem[{Chen et~al.(2024)Chen, Wang, Cao, Liu, Gao, Cui, Zhu, Ye, Tian, Liu et~al.}]{chen2024expanding}
Zhe Chen, Weiyun Wang, Yue Cao, Yangzhou Liu, Zhangwei Gao, Erfei Cui, Jinguo Zhu, Shenglong Ye, Hao Tian, Zhaoyang Liu, and 1 others. 2024.
\newblock Expanding performance boundaries of open-source multimodal models with model, data, and test-time scaling.
\newblock \emph{arXiv preprint arXiv:2412.05271}.

\bibitem[{Cheng et~al.(2024)Cheng, Yin, Fu, Guo, Yang, Kautz, Wang, and Liu}]{cheng_spatialrgpt_2024}
An-Chieh Cheng, Hongxu Yin, Yang Fu, Qiushan Guo, Ruihan Yang, Jan Kautz, Xiaolong Wang, and Sifei Liu. 2024.
\newblock \href {https://doi.org/10.48550/arXiv.2406.01584} {{SpatialRGPT}: Grounded spatial reasoning in vision language models}.
\newblock \emph{Preprint}, arxiv:2406.01584 [cs].
\newblock Version: 3.

\bibitem[{{Claude Team}(2024)}]{claudesonnet35}
{Claude Team}. 2024.
\newblock Introducing the next generation of claude.
\newblock \url{https://www.anthropic.com/news/claude-3-family}.

\bibitem[{Dosovitskiy et~al.(2020)Dosovitskiy, Beyer, Kolesnikov, Weissenborn, Zhai, Unterthiner, Dehghani, Minderer, Heigold, Gelly, Uszkoreit, and Houlsby}]{Dosovitskiy2020AnII}
Alexey Dosovitskiy, Lucas Beyer, Alexander Kolesnikov, Dirk Weissenborn, Xiaohua Zhai, Thomas Unterthiner, Mostafa Dehghani, Matthias Minderer, Georg Heigold, Sylvain Gelly, Jakob Uszkoreit, and Neil Houlsby. 2020.
\newblock \href {https://api.semanticscholar.org/CorpusID:225039882} {An image is worth 16x16 words: Transformers for image recognition at scale}.
\newblock \emph{ArXiv}, abs/2010.11929.

\bibitem[{Duan et~al.(2025)Duan, Xiu, and Gorlatova}]{duan_advancing_2025}
Lin Duan, Yanming Xiu, and Maria Gorlatova. 2025.
\newblock \href {https://doi.org/10.48550/arXiv.2501.13964} {Advancing the understanding and evaluation of {AR}-generated scenes: When vision-language models shine and stumble}.
\newblock \emph{Preprint}, arxiv:2501.13964 [cs].

\bibitem[{Estermann et~al.(2024)Estermann, Lanzendörfer, Niedermayr, and Wattenhofer}]{estermann_puzzles_2024}
Benjamin Estermann, Luca~A. Lanzendörfer, Yannick Niedermayr, and Roger Wattenhofer. 2024.
\newblock \href {https://doi.org/10.48550/arXiv.2407.00401} {{PUZZLES}: A benchmark for neural algorithmic reasoning}.
\newblock \emph{Preprint}, arxiv:2407.00401 [cs].

\bibitem[{Feng et~al.(2025)Feng, Han, Yang, Yue, Levine, and Luo}]{feng_reflective_2025}
Yunhai Feng, Jiaming Han, Zhuoran Yang, Xiangyu Yue, Sergey Levine, and Jianlan Luo. 2025.
\newblock \href {https://doi.org/10.48550/arXiv.2502.16707} {Reflective planning: Vision-language models for multi-stage long-horizon robotic manipulation}.
\newblock \emph{Preprint}, arxiv:2502.16707 [cs].

\bibitem[{{Gemini Team}(2024)}]{gemini20flash}
{Gemini Team}. 2024.
\newblock Gemini 2.0 flash (experimental).

\bibitem[{Gozon and Yu(2024)}]{gozon2024computing}
Marcus Gozon and Jingjin Yu. 2024.
\newblock On computing makespan-optimal solutions for generalized sliding-tile puzzles.
\newblock In \emph{Proceedings of the AAAI Conference on Artificial Intelligence}, volume 38(9), pages 10288--10296.

\bibitem[{Guruprasad et~al.(2024)Guruprasad, Sikka, Song, Wang, and Liang}]{guruprasad_benchmarking_2024}
Pranav Guruprasad, Harshvardhan Sikka, Jaewoo Song, Yangyue Wang, and Paul~Pu Liang. 2024.
\newblock \href {https://doi.org/10.48550/arXiv.2411.05821} {Benchmarking vision, language, \& action models on robotic learning tasks}.
\newblock \emph{Preprint}, arxiv:2411.05821 [cs].

\bibitem[{Hart et~al.(1968)Hart, Nilsson, and Raphael}]{hart1968formal}
Peter~E Hart, Nils~J Nilsson, and Bertram Raphael. 1968.
\newblock A formal basis for the heuristic determination of minimum cost paths.
\newblock \emph{IEEE transactions on Systems Science and Cybernetics}, 4(2):100--107.

\bibitem[{Hu et~al.(2023)Hu, Lin, Zhang, Yi, and Gao}]{hu_look_2023}
Yingdong Hu, Fanqi Lin, Tong Zhang, Li~Yi, and Yang Gao. 2023.
\newblock \href {https://doi.org/10.48550/arXiv.2311.17842} {Look before you leap: Unveiling the power of {GPT}-4v in robotic vision-language planning}.
\newblock \emph{Preprint}, arxiv:2311.17842 [cs].

\bibitem[{Ilharco et~al.(2021)Ilharco, Zellers, Farhadi, and Hajishirzi}]{ilharco_probing_2021}
Gabriel Ilharco, Rowan Zellers, Ali Farhadi, and Hannaneh Hajishirzi. 2021.
\newblock \href {https://doi.org/10.48550/arXiv.2005.00619} {Probing contextual language models for common ground with visual representations}.
\newblock \emph{Preprint}, arxiv:2005.00619 [cs].

\bibitem[{Jassim et~al.(2024)Jassim, Holubar, Richter, Wolff, Ohmer, and Bruni}]{jassim_grasp_2024}
Serwan Jassim, Mario Holubar, Annika Richter, Cornelius Wolff, Xenia Ohmer, and Elia Bruni. 2024.
\newblock \href {https://doi.org/10.48550/arXiv.2311.09048} {{GRASP}: A novel benchmark for evaluating language {GRounding} and situated physics understanding in multimodal language models}.
\newblock \emph{Preprint}, arxiv:2311.09048.

\bibitem[{Johnson et~al.(2016)Johnson, Hariharan, Maaten, Fei-Fei, Zitnick, and Girshick}]{johnson_clevr_2016}
Justin Johnson, Bharath Hariharan, Laurens van~der Maaten, Li~Fei-Fei, C.~Lawrence Zitnick, and Ross Girshick. 2016.
\newblock \href {https://doi.org/10.48550/arXiv.1612.06890} {{CLEVR}: A diagnostic dataset for compositional language and elementary visual reasoning}.
\newblock \emph{Preprint}, arxiv:1612.06890 [cs].

\bibitem[{Kamath et~al.(2023{\natexlab{a}})Kamath, Hessel, and Chang}]{DBLP:conf/emnlp/KamathHC23a}
Amita Kamath, Jack Hessel, and Kai{-}Wei Chang. 2023{\natexlab{a}}.
\newblock \href {https://doi.org/10.18653/V1/2023.EMNLP-MAIN.568} {What's "up" with vision-language models? investigating their struggle with spatial reasoning}.
\newblock In \emph{Proceedings of the 2023 Conference on Empirical Methods in Natural Language Processing, {EMNLP} 2023, Singapore, December 6-10, 2023}, pages 9161--9175. Association for Computational Linguistics.

\bibitem[{Kamath et~al.(2023{\natexlab{b}})Kamath, Hessel, and Chang}]{kamath_whats_2023}
Amita Kamath, Jack Hessel, and Kai-Wei Chang. 2023{\natexlab{b}}.
\newblock \href {https://doi.org/10.48550/arXiv.2310.19785} {What's "up" with vision-language models? investigating their struggle with spatial reasoning}.
\newblock \emph{Preprint}, arxiv:2310.19785 [cs].

\bibitem[{Kuhnle and Copestake(2017)}]{kuhnle_shapeworld_2017}
Alexander Kuhnle and Ann Copestake. 2017.
\newblock \href {https://doi.org/10.48550/arXiv.1704.04517} {{ShapeWorld} - a new test methodology for multimodal language understanding}.
\newblock \emph{Preprint}, arxiv:1704.04517 [cs].

\bibitem[{Li et~al.(2024{\natexlab{a}})Li, Zhang, Guo, Zhang, Li, Zhang, Zhang, Li, Liu, and Li}]{li2024llava}
Bo~Li, Yuanhan Zhang, Dong Guo, Renrui Zhang, Feng Li, Hao Zhang, Kaichen Zhang, Yanwei Li, Ziwei Liu, and Chunyuan Li. 2024{\natexlab{a}}.
\newblock Llava-onevision: Easy visual task transfer.
\newblock \emph{arXiv preprint arXiv:2408.03326}.

\bibitem[{Li et~al.(2024{\natexlab{b}})Li, Yu, Zheng, Zhan, Zhang, Zhang, Yang, Chen, Sun, Cao, Shen, Li, Tao, and He}]{li_muep_2024}
Kanxue Li, Baosheng Yu, Qi~Zheng, Yibing Zhan, Yuhui Zhang, Tianle Zhang, Yijun Yang, Yue Chen, Lei Sun, Qiong Cao, Li~Shen, Lusong Li, Dapeng Tao, and Xiaodong He. 2024{\natexlab{b}}.
\newblock \href {https://doi.org/10.24963/ijcai.2024/15} {{MuEP}: A multimodal benchmark for embodied planning with foundation models}.
\newblock In \emph{Proceedings of the Thirty-{ThirdInternational} Joint Conference on Artificial Intelligence}, pages 129--138. International Joint Conferences on Artificial Intelligence Organization.

\bibitem[{Li et~al.(2023)Li, Wang, Stengel-Eskin, Kortylewski, Ma, Durme, and Yuille}]{li_super-clevr_2023}
Zhuowan Li, Xingrui Wang, Elias Stengel-Eskin, Adam Kortylewski, Wufei Ma, Benjamin~Van Durme, and Alan Yuille. 2023.
\newblock \href {https://doi.org/10.48550/arXiv.2212.00259} {Super-{CLEVR}: A virtual benchmark to diagnose domain robustness in visual reasoning}.
\newblock \emph{Preprint}, arxiv:2212.00259 [cs].

\bibitem[{Liu et~al.(2023)Liu, Emerson, and Collier}]{liu_visual_2023}
Fangyu Liu, Guy Emerson, and Nigel Collier. 2023.
\newblock \href {https://doi.org/10.48550/arXiv.2205.00363} {Visual spatial reasoning}.
\newblock \emph{Preprint}, arxiv:2205.00363 [cs].

\bibitem[{Mecattaf et~al.(2024)Mecattaf, Slater, Tešić, Prunty, Voudouris, and Cheke}]{mecattaf_little_2024}
Matteo~G. Mecattaf, Ben Slater, Marko Tešić, Jonathan Prunty, Konstantinos Voudouris, and Lucy~G. Cheke. 2024.
\newblock \href {https://arxiv.org/abs/2410.23242 [cs]} {A little less conversation, a little more action, please: Investigating the physical common-sense of {LLMs} in a 3d embodied environment}.
\newblock \emph{Preprint}, arxiv:2410.23242 [cs].

\bibitem[{Merullo et~al.(2023)Merullo, Castricato, Eickhoff, and Pavlick}]{merullo_linearly_2023}
Jack Merullo, Louis Castricato, Carsten Eickhoff, and Ellie Pavlick. 2023.
\newblock \href {https://doi.org/10.48550/arXiv.2209.15162} {Linearly mapping from image to text space}.
\newblock \emph{Preprint}, arxiv:2209.15162 [cs].

\bibitem[{Mirzaee et~al.(2021)Mirzaee, Faghihi, Ning, and Kordjmashidi}]{mirzaee_spartqa_2021}
Roshanak Mirzaee, Hossein~Rajaby Faghihi, Qiang Ning, and Parisa Kordjmashidi. 2021.
\newblock \href {https://doi.org/10.48550/arXiv.2104.05832} {{SpartQA}: : A textual question answering benchmark for spatial reasoning}.
\newblock \emph{Preprint}, arxiv:2104.05832 [cs].

\bibitem[{Oliveira et~al.(2024)Oliveira, Brandão, Luz, Martins, Soares, and Melo}]{oliveira_sliding_2024}
Bryan Lincoln Marques~de Oliveira, Bruno Brandão, Murilo Lopes~da Luz, Luana Guedes~Barros Martins, Telma Woerle de~Lima Soares, and Luckeciano~Carvalho Melo. 2024.
\newblock \href {https://openreview.net/forum?id=VHBkmgWE0o} {Sliding puzzles gym: A scalable benchmark for state representation in visual reinforcement learning}.
\newblock In \emph{{NeurIPS} 2024 Workshop on Open-World Agents}.

\bibitem[{OpenAI et~al.(2024)OpenAI, Hurst, Lerer, Goucher, Perelman, Ramesh, Clark, Ostrow, Welihinda, Hayes, Radford, Mądry, Baker-Whitcomb, Beutel, Borzunov, Carney, Chow, Kirillov, Nichol, Paino, Renzin, Passos, Kirillov, Christakis, Conneau, Kamali, Jabri, Moyer, Tam, Crookes, Tootoochian, Tootoonchian, Kumar, Vallone, Karpathy, Braunstein, Cann, Codispoti, Galu, Kondrich, Tulloch, Mishchenko, Baek, Jiang, Pelisse, Woodford, Gosalia, Dhar, Pantuliano, Nayak, Oliver, Zoph, Ghorbani, Leimberger, Rossen, Sokolowsky, Wang, Zweig, Hoover, Samic, McGrew, Spero, Giertler, Cheng, Lightcap, Walkin, Quinn, Guarraci, Hsu, Kellogg, Eastman, Lugaresi, Wainwright, Bassin, Hudson, Chu, Nelson, Li, Shern, Conger, Barette, Voss, Ding, Lu, Zhang, Beaumont, Hallacy, Koch, Gibson, Kim, Choi, McLeavey, Hesse, Fischer, Winter, Czarnecki, Jarvis, Wei, Koumouzelis, Sherburn, Kappler, Levin, Levy, Carr, Farhi, Mely, Robinson, Sasaki, Jin, Valladares, Tsipras, Li, Nguyen, Findlay, Oiwoh, Wong, Asdar, Proehl, Yang, Antonow,
  Kramer, Peterson, Sigler, Wallace, Brevdo, Mays, Khorasani, Such, Raso, Zhang, von Lohmann, Sulit, Goh, Oden, Salmon, Starace, Brockman, Salman, Bao, Hu, Wong, Wang, Schmidt, Whitney, Jun, Kirchner, de~Oliveira~Pinto, Ren, Chang, Chung, Kivlichan, O'Connell, O'Connell, Osband, Silber, Sohl, Okuyucu, Lan, Kostrikov, Sutskever, Kanitscheider, Gulrajani, Coxon, Menick, Pachocki, Aung, Betker, Crooks, Lennon, Kiros, Leike, Park, Kwon, Phang, Teplitz, Wei, Wolfe, Chen, Harris, Varavva, Lee, Shieh, Lin, Yu, Weng, Tang, Yu, Jang, Candela, Beutler, Landers, Parish, Heidecke, Schulman, Lachman, McKay, Uesato, Ward, Kim, Huizinga, Sitkin, Kraaijeveld, Gross, Kaplan, Snyder, Achiam, Jiao, Lee, Zhuang, Harriman, Fricke, Hayashi, Singhal, Shi, Karthik, Wood, Rimbach, Hsu, Nguyen, Gu-Lemberg, Button, Liu, Howe, Muthukumar, Luther, Ahmad, Kai, Itow, Workman, Pathak, Chen, Jing, Guy, Fedus, Zhou, Mamitsuka, Weng, McCallum, Held, Ouyang, Feuvrier, Zhang, Kondraciuk, Kaiser, Hewitt, Metz, Doshi, Aflak, Simens, Boyd,
  Thompson, Dukhan, Chen, Gray, Hudnall, Zhang, Aljubeh, Litwin, Zeng, Johnson, Shetty, Gupta, Shah, Yatbaz, Yang, Zhong, Glaese, Chen, Janner, Lampe, Petrov, Wu, Wang, Fradin, Pokrass, Castro, de~Castro, Pavlov, Brundage, Wang, Khan, Murati, Bavarian, Lin, Yesildal, Soto, Gimelshein, Cone, Staudacher, Summers, LaFontaine, Chowdhury, Ryder, Stathas, Turley, Tezak, Felix, Kudige, Keskar, Deutsch, Bundick, Puckett, Nachum, Okelola, Boiko, Murk, Jaffe, Watkins, Godement, Campbell-Moore, Chao, McMillan, Belov, Su, Bak, Bakkum, Deng, Dolan, Hoeschele, Welinder, Tillet, Pronin, Tillet, Dhariwal, Yuan, Dias, Lim, Arora, Troll, Lin, Lopes, Puri, Miyara, Leike, Gaubert, Zamani, Wang, Donnelly, Honsby, Smith, Sahai, Ramchandani, Huet, Carmichael, Zellers, Chen, Chen, Nigmatullin, Cheu, Jain, Altman, Schoenholz, Toizer, Miserendino, Agarwal, Culver, Ethersmith, Gray, Grove, Metzger, Hermani, Jain, Zhao, Wu, Jomoto, Wu, Shuaiqi, Xia, Phene, Papay, Narayanan, Coffey, Lee, Hall, Balaji, Broda, Stramer, Xu, Gogineni,
  Christianson, Sanders, Patwardhan, Cunninghman, Degry, Dimson, Raoux, Shadwell, Zheng, Underwood, Markov, Sherbakov, Rubin, Stasi, Kaftan, Heywood, Peterson, Walters, Eloundou, Qi, Moeller, Monaco, Kuo, Fomenko, Chang, Zheng, Zhou, Manassra, Sheu, Zaremba, Patil, Qian, Kim, Cheng, Zhang, He, Zhang, Jin, Dai, and Malkov}]{gpt4o}
OpenAI, Aaron Hurst, Adam Lerer, Adam~P. Goucher, Adam Perelman, Aditya Ramesh, Aidan Clark, AJ~Ostrow, Akila Welihinda, Alan Hayes, Alec Radford, Aleksander Mądry, Alex Baker-Whitcomb, Alex Beutel, Alex Borzunov, Alex Carney, Alex Chow, Alex Kirillov, Alex Nichol, and 400 others. 2024.
\newblock \href {https://arxiv.org/abs/2410.21276} {Gpt-4o system card}.
\newblock \emph{Preprint}, arXiv:2410.21276.

\bibitem[{Patel and Pavlick(2021)}]{patel_mapping_2021}
Roma Patel and Ellie Pavlick. 2021.
\newblock \href {https://openreview.net/forum?id=gJcEM8sxHK} {Mapping language models to grounded conceptual spaces}.
\newblock In \emph{International Conference on Learning Representations}.

\bibitem[{Rajabi and Kosecka(2024{\natexlab{a}})}]{rajabi_gsr-bench_2024}
Navid Rajabi and Jana Kosecka. 2024{\natexlab{a}}.
\newblock \href {https://doi.org/10.48550/arXiv.2406.13246} {{GSR}-{BENCH}: A benchmark for grounded spatial reasoning evaluation via multimodal {LLMs}}.
\newblock \emph{Preprint}, arxiv:2406.13246 [cs].
\newblock Version: 2.

\bibitem[{Rajabi and Kosecka(2024{\natexlab{b}})}]{rajabi_towards_2024}
Navid Rajabi and Jana Kosecka. 2024{\natexlab{b}}.
\newblock \href {https://doi.org/10.48550/arXiv.2308.09778} {Towards grounded visual spatial reasoning in multi-modal vision language models}.
\newblock \emph{Preprint}, arxiv:2308.09778 [cs].

\bibitem[{Rizvi et~al.(2024)Rizvi, Zhu, and Gurevych}]{rizvi_sparc_2024}
Md~Imbesat~Hassan Rizvi, Xiaodan Zhu, and Iryna Gurevych. 2024.
\newblock \href {https://doi.org/10.48550/arXiv.2406.04566} {{SpaRC} and {SpaRP}: Spatial reasoning characterization and path generation for understanding spatial reasoning capability of large language models}.
\newblock \emph{Preprint}, arxiv:2406.04566 [cs].

\bibitem[{Roberts and Roberts(2024)}]{roberts_smart_2024}
Denisa Roberts and Lucas Roberts. 2024.
\newblock \href {https://doi.org/10.48550/arXiv.2407.04212} {Smart vision-language reasoners}.
\newblock \emph{Preprint}, arxiv:2407.04212 [cs].

\bibitem[{Su et~al.(2024)Su, Ling, Shi, Cheng, Yim, and Song}]{su_actplan-1k_2024}
Ying Su, Zhan Ling, Haochen Shi, Jiayang Cheng, Yauwai Yim, and Yangqiu Song. 2024.
\newblock \href {https://doi.org/10.48550/arXiv.2410.03907} {{ActPlan}-1k: Benchmarking the procedural planning ability of visual language models in household activities}.
\newblock \emph{Preprint}, arxiv:2410.03907 [cs].

\bibitem[{Tang et~al.(2024)Tang, Qu, Wang, Zhuang, Wu, Ma, Wang, Zheng, Zhao, and Zhao}]{tang_sparkle_2024}
Yihong Tang, Ao~Qu, Zhaokai Wang, Dingyi Zhuang, Zhaofeng Wu, Wei Ma, Shenhao Wang, Yunhan Zheng, Zhan Zhao, and Jinhua Zhao. 2024.
\newblock \href {https://doi.org/10.48550/arXiv.2410.16162} {Sparkle: Mastering basic spatial capabilities in vision language models elicits generalization to composite spatial reasoning}.
\newblock \emph{Preprint}, arxiv:2410.16162 [cs].

\bibitem[{Wang et~al.(2024{\natexlab{a}})Wang, Ming, Shi, Vineet, Wang, and Joshi}]{DBLP:journals/corr/abs-2406-14852}
Jiayu Wang, Yifei Ming, Zhenmei Shi, Vibhav Vineet, Xin Wang, and Neel Joshi. 2024{\natexlab{a}}.
\newblock \href {https://doi.org/10.48550/ARXIV.2406.14852} {Is {A} picture worth {A} thousand words? delving into spatial reasoning for vision language models}.
\newblock \emph{CoRR}, abs/2406.14852.

\bibitem[{Wang et~al.(2024{\natexlab{b}})Wang, Ming, Shi, Vineet, Wang, Li, and Joshi}]{wang_is_2024}
Jiayu Wang, Yifei Ming, Zhenmei Shi, Vibhav Vineet, Xin Wang, Yixuan Li, and Neel Joshi. 2024{\natexlab{b}}.
\newblock \href {https://doi.org/10.48550/arXiv.2406.14852} {Is a picture worth a thousand words? delving into spatial reasoning for vision language models}.
\newblock \emph{Preprint}, arxiv:2406.14852 [cs].

\bibitem[{Wang et~al.(2024{\natexlab{c}})Wang, Zhang, Unjhawala, Negrut, Wang, Slaton, Serban, Wu, and Negrut}]{wang_simbench_2024}
Jingquan Wang, Harry Zhang, Huzaifa~Mustafa Unjhawala, Peter Negrut, Shu Wang, Khailanii Slaton, Radu Serban, Jin-Long Wu, and Dan Negrut. 2024{\natexlab{c}}.
\newblock \href {https://arxiv.org/abs/2408.11987 [cs]} {{SimBench}: A rule-based multi-turn interaction benchmark for evaluating an {LLM}'s ability to generate digital twins}.
\newblock \emph{Preprint}, arxiv:2408.11987 [cs].

\bibitem[{Wang et~al.(2024{\natexlab{d}})Wang, Bai, Tan, Wang, Fan, Bai, Chen, Liu, Wang, Ge, Fan, Dang, Du, Ren, Men, Liu, Zhou, Zhou, and Lin}]{Qwen2VL}
Peng Wang, Shuai Bai, Sinan Tan, Shijie Wang, Zhihao Fan, Jinze Bai, Keqin Chen, Xuejing Liu, Jialin Wang, Wenbin Ge, Yang Fan, Kai Dang, Mengfei Du, Xuancheng Ren, Rui Men, Dayiheng Liu, Chang Zhou, Jingren Zhou, and Junyang Lin. 2024{\natexlab{d}}.
\newblock Qwen2-vl: Enhancing vision-language model's perception of the world at any resolution.
\newblock \emph{arXiv preprint arXiv:2409.12191}.

\bibitem[{Wang et~al.(2016)Wang, Liang, and Manning}]{wang_learning_2016}
Sida~I. Wang, Percy Liang, and Christopher~D. Manning. 2016.
\newblock \href {https://arxiv.org/abs/1606.02447 [cs]} {Learning language games through interaction}.
\newblock \emph{Preprint}, arxiv:1606.02447 [cs].

\bibitem[{Wang et~al.(2023)Wang, Ma, Li, Kortylewski, and Yuille}]{wang_3d_aware_2023}
Xingrui Wang, Wufei Ma, Zhuowan Li, Adam Kortylewski, and Alan Yuille. 2023.
\newblock \href {https://doi.org/10.48550/arXiv.2310.17914} {3d-aware visual question answering about parts, poses and occlusions}.
\newblock \emph{arXiv preprint}.

\bibitem[{Wang et~al.(2025)Wang, Zhuang, and Wu}]{wang_are_2025}
Xinyu Wang, Bohan Zhuang, and Qi~Wu. 2025.
\newblock \href {https://doi.org/10.48550/arXiv.2503.02358} {Are large vision language models good game players?}
\newblock \emph{Preprint}, arxiv:2503.02358 [cs].

\bibitem[{Wei et~al.(2022)Wei, Wang, Schuurmans, Bosma, Xia, Chi, Le, Zhou et~al.}]{wei2022chain}
Jason Wei, Xuezhi Wang, Dale Schuurmans, Maarten Bosma, Fei Xia, Ed~Chi, Quoc~V Le, Denny Zhou, and 1 others. 2022.
\newblock Chain-of-thought prompting elicits reasoning in large language models.
\newblock \emph{Advances in neural information processing systems}, 35:24824--24837.

\bibitem[{Wu et~al.(2024)Wu, Zhao, Saxon, Bui, Wang, Zhang, and Chang}]{wu_vsp_2024}
Qiucheng Wu, Handong Zhao, Michael Saxon, Trung Bui, William~Yang Wang, Yang Zhang, and Shiyu Chang. 2024.
\newblock \href {https://doi.org/10.48550/arXiv.2407.01863} {{VSP}: Assessing the dual challenges of perception and reasoning in spatial planning tasks for {VLMs}}.
\newblock \emph{Preprint}, arxiv:2407.01863 [cs].

\bibitem[{Xi et~al.(2023)Xi, Chen, Guo, He, Ding, Hong, Zhang, Wang, Jin, Zhou et~al.}]{xi2023rise}
Zhiheng Xi, Wenxiang Chen, Xin Guo, Wei He, Yiwen Ding, Boyang Hong, Ming Zhang, Junzhe Wang, Senjie Jin, Enyu Zhou, and 1 others. 2023.
\newblock The rise and potential of large language model based agents: A survey.
\newblock \emph{arXiv preprint arXiv:2309.07864}.

\bibitem[{Yamada et~al.(2024)Yamada, Bao, Lampinen, Kasai, and Yildirim}]{yamada_evaluating_2024}
Yutaro Yamada, Yihan Bao, Andrew~K. Lampinen, Jungo Kasai, and Ilker Yildirim. 2024.
\newblock \href {https://doi.org/10.48550/arXiv.2310.14540} {Evaluating spatial understanding of large language models}.
\newblock \emph{Preprint}, arxiv:2310.14540 [cs].

\bibitem[{Yang et~al.(2024{\natexlab{a}})Yang, Yang, Hui, Zheng, Yu, Zhou, Li, Li, Liu, Huang, Dong, Wei, Lin, Tang, Wang, Yang, Tu, Zhang, Ma, Xu, Zhou, Bai, He, Lin, Dang, Lu, Chen, Yang, Li, Xue, Ni, Zhang, Wang, Peng, Men, Gao, Lin, Wang, Bai, Tan, Zhu, Li, Liu, Ge, Deng, Zhou, Ren, Zhang, Wei, Ren, Fan, Yao, Zhang, Wan, Chu, Liu, Cui, Zhang, and Fan}]{qwen2}
An~Yang, Baosong Yang, Binyuan Hui, Bo~Zheng, Bowen Yu, Chang Zhou, Chengpeng Li, Chengyuan Li, Dayiheng Liu, Fei Huang, Guanting Dong, Haoran Wei, Huan Lin, Jialong Tang, Jialin Wang, Jian Yang, Jianhong Tu, Jianwei Zhang, Jianxin Ma, and 40 others. 2024{\natexlab{a}}.
\newblock Qwen2 technical report.
\newblock \emph{arXiv preprint arXiv:2407.10671}.

\bibitem[{Yang et~al.(2024{\natexlab{b}})Yang, Yang, Zhang, Hui, Zheng, Yu, Li, Liu, Huang, Wei, Lin, Yang, Tu, Zhang, Yang, Yang, Zhou, Lin, Dang, Lu, Bao, Yang, Yu, Li, Xue, Zhang, Zhu, Men, Lin, Li, Xia, Ren, Ren, Fan, Su, Zhang, Wan, Liu, Cui, Zhang, and Qiu}]{qwen2.5}
An~Yang, Baosong Yang, Beichen Zhang, Binyuan Hui, Bo~Zheng, Bowen Yu, Chengyuan Li, Dayiheng Liu, Fei Huang, Haoran Wei, Huan Lin, Jian Yang, Jianhong Tu, Jianwei Zhang, Jianxin Yang, Jiaxi Yang, Jingren Zhou, Junyang Lin, Kai Dang, and 22 others. 2024{\natexlab{b}}.
\newblock Qwen2.5 technical report.
\newblock \emph{arXiv preprint arXiv:2412.15115}.

\bibitem[{Zeng et~al.(2023)Zeng, Gan, Wang, Liu, and Yu}]{zeng2023large}
Fanlong Zeng, Wensheng Gan, Yongheng Wang, Ning Liu, and Philip~S Yu. 2023.
\newblock Large language models for robotics: A survey.
\newblock \emph{arXiv preprint arXiv:2311.07226}.

\bibitem[{Zhai et~al.(2023)Zhai, Mustafa, Kolesnikov, and Beyer}]{zhai2023sigmoid}
Xiaohua Zhai, Basil Mustafa, Alexander Kolesnikov, and Lucas Beyer. 2023.
\newblock \href {https://arxiv.org/abs/2303.15343} {Sigmoid loss for language image pre-training}.
\newblock \emph{Preprint}, arXiv:2303.15343.

\bibitem[{Zhang et~al.(2024)Zhang, Cheng, He, Wang, Shen, Tan, Hou, He, Ma, Lu, and Zhuang}]{zhang_multimodal_2024}
Wenqi Zhang, Zhenglin Cheng, Yuanyu He, Mengna Wang, Yongliang Shen, Zeqi Tan, Guiyang Hou, Mingqian He, Yanna Ma, Weiming Lu, and Yueting Zhuang. 2024.
\newblock \href {https://doi.org/10.48550/arXiv.2407.07053} {Multimodal self-instruct: Synthetic abstract image and visual reasoning instruction using language model}.
\newblock \emph{Preprint}, arxiv:2407.07053 [cs].

\bibitem[{Zhao et~al.(2025)Zhao, Wang, Fang, Gao, Man, Cui, Wang, Chen, Li, and Zhu}]{zhao_embodied-r_2025}
Baining Zhao, Ziyou Wang, Jianjie Fang, Chen Gao, Fanhang Man, Jinqiang Cui, Xin Wang, Xinlei Chen, Yong Li, and Wenwu Zhu. 2025.
\newblock \href {https://doi.org/10.48550/arXiv.2504.12680} {Embodied-r: Collaborative framework for activating embodied spatial reasoning in foundation models via reinforcement learning}.
\newblock \emph{Preprint}, arxiv:2504.12680 [cs].

\end{thebibliography}
\end{document}